\documentclass[nonacm,sigplan]{acmart}

\usepackage[]{hyperref}

\usepackage{multirow}
\usepackage{xspace}
\usepackage{tikz}
\usepackage{array}
\usepackage{listings}
\usepackage{subcaption}
\usepackage{lipsum}

\newcommand{\name}{GS-Scale\xspace}

\newcommand*\circled[1]{\tikz[baseline=(char.base)]{            \node[shape=circle,fill,inner sep=1pt] (char) {\textcolor{white}{#1}};}}

\definecolor{mauvelous}{rgb}{0.94, 0.6, 0.67}
\newcommand{\minus}{\scalebox{0.75}[1.0]{$-$}}

\lstdefinestyle{myStyle1}{
  belowcaptionskip=1\baselineskip,
  frame=tb,
  language=c++,
  aboveskip=0mm,
  belowskip=0mm,
  showstringspaces=false,
  columns=flexible,
  basicstyle={\fontsize{8.0pt}{8.0pt}\fontfamily{pcr}\selectfont},
  numbers=left,
  xleftmargin=2em,
  numberstyle={\color{gray}\texttt},
  keywordstyle=\color{black}\textbf,
  commentstyle=\color{mauvelous},
  stringstyle=\color{mauvelous},
  frame=none,
  breaklines=true,
  breakatwhitespace=true,
  tabsize=2,
  morekeywords={WHILE, IF, FOR, ELSE, IS, IN, TO, FROM},
  deletekeywords={and, sizeof},
}
\lstdefinestyle{myStyle2}{
  belowcaptionskip=1\baselineskip,
  frame=tb,
  language=c++,
  aboveskip=0mm,
  belowskip=0mm,
  showstringspaces=false,
  columns=flexible,
  basicstyle={\fontsize{6.6pt}{6.6pt}\fontfamily{fvm}\selectfont},
  numbers=left,
  xleftmargin=2.5em,
  numberstyle={\color{gray}\texttt},
  keywordstyle=\color{black}\textbf,
  commentstyle=\color{mauvelous}\textbf,
  stringstyle=\color{mauve},
  frame=none,
  breaklines=true,
  breakatwhitespace=true,
  tabsize=2,
  morekeywords={MatrixMult, Softmax, parfor, parallel, each, not, in, intersection, map, max, erase},
  deletekeywords={and},
  moredelim=**[is][\color{red}]{@}{@},
}

\author{
Donghyun Lee$^{1}$ \quad Dawoon Jeong$^{1}$ \quad Jae W. Lee$^{1}$ \quad Hongil Yoon$^{1,2}$\\
$^{1}$Seoul National University \quad
$^{2}$Google\\
{\tt\small \{eudh1206, daun20211, jaewlee\}@snu.ac.kr, hongilyoon@google.com}
}

\renewcommand{\shortauthors}{Lee, et al.}

\begin{document}

\title{\name: Unlocking Large-Scale 3D Gaussian Splatting Training via Host Offloading}


\begin{abstract}

The advent of 3D Gaussian Splatting has revolutionized graphics rendering by delivering high visual quality and fast rendering speeds. However, training large-scale scenes at high quality remains challenging due to the substantial memory demands required to store parameters, gradients, and optimizer states, which can quickly overwhelm GPU memory.
To address these limitations, we propose \name, a fast and memory-efficient training system for 3D Gaussian Splatting. \name stores all Gaussians in host memory, transferring only a subset to the GPU on demand for each forward and backward pass. While this dramatically reduces GPU memory usage, it requires frustum culling and optimizer updates to be executed on the CPU, introducing slowdowns due to CPU's limited compute and memory bandwidth.
To mitigate this, \name employs three system-level optimizations: (1) \textit{selective offloading} of geometric parameters for fast frustum culling, (2) \textit{parameter forwarding} to pipeline CPU optimizer updates with GPU computation, and (3) \textit{deferred optimizer update} to minimize unnecessary memory accesses for Gaussians with zero gradients. Our extensive evaluations on large-scale datasets demonstrate that \name significantly lowers GPU memory demands by 3.3-5.6$\times$, while achieving training speeds comparable to GPU without host offloading. This enables large-scale 3D Gaussian Splatting training on consumer-grade GPUs; for instance, \name can scale the number of Gaussians from 4 million to 18 million on an RTX 4070 Mobile GPU, leading to 23-35\% LPIPS (learned perceptual image patch similarity) improvement.

\end{abstract}

\maketitle

\section{Introduction}
\label{sec:intro}

\begin{figure}[t]
  \centering
  \includegraphics[width=\columnwidth]{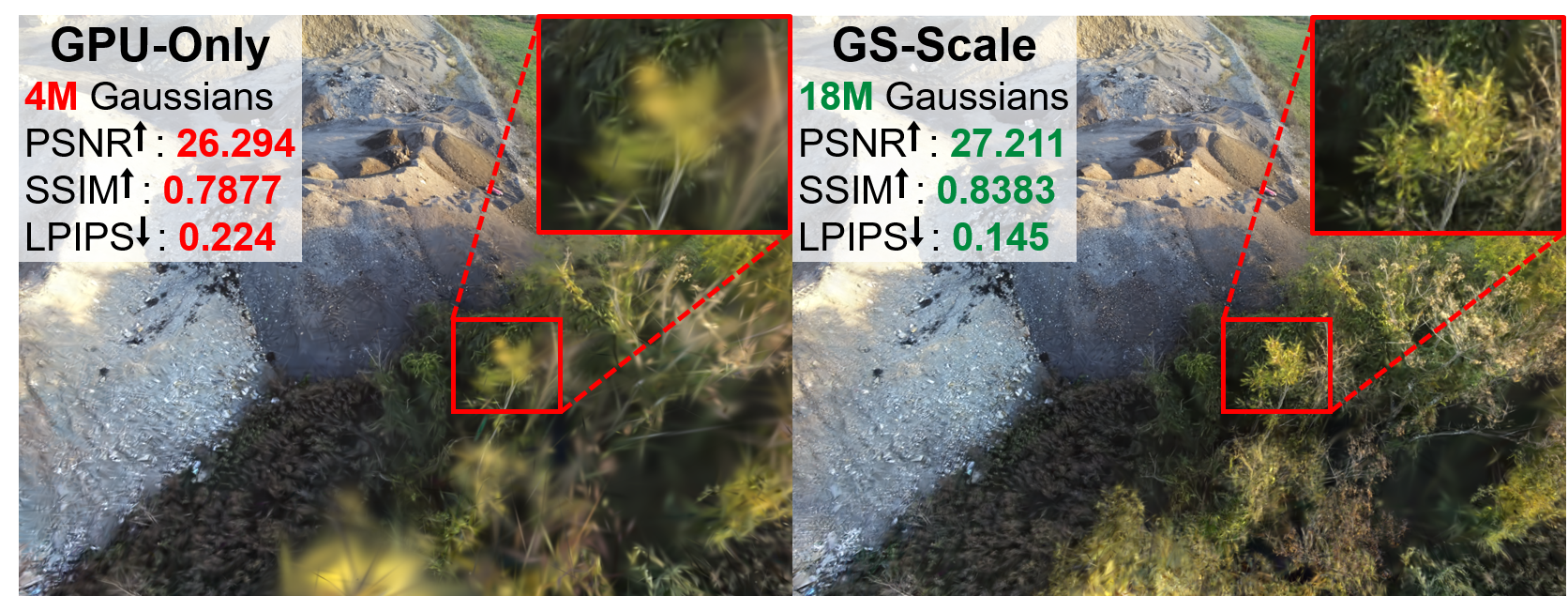}
  \caption{Comparison of the maximum rendering quality achievable in 3DGS training using a GPU-only system and \name. Training is conducted on RTX 4070 Mobile GPU with Rubble scene. Higher is better for PSNR and SSIM, lower is better for LPIPS.}
  \label{fig:thumbnail}
\end{figure}

Differentiable rendering methods~\cite{3dgs, nerf, instantngp, plenoxels} have significantly improved the quality and efficiency of novel view synthesis. Among these innovations, 3D Gaussian Splatting (3DGS)~\cite{3dgs} has emerged as a state-of-the-art technique, offering high visual quality and fast rendering by representing a 3D scene with millions of trainable 3D Gaussian primitives. 

However, the increasing demand for reconstructing larger and more visually detailed 3D scenes has led to a significant surge in the number of Gaussians required during training~\cite{grendel, citygaussian, citygaussianv2, vastgaussian, hug, retinags}, pushing the limit of GPU memory. For example, in Rubble~\cite{mega-nerf} scene, reaching the highest visual quality requires about 40 million Gaussians resulting in 53 GB of GPU memory, far exceeding the capacity of any single consumer-grade GPU.
These high memory demands present a major obstacle to scaling the number of Gaussians in 3DGS training, leading to reduced scene expressiveness and, consequently, degraded rendering quality. 

Recent works~\cite{grendel, retinags, vastgaussian} have addressed these challenges through distributed training across multiple GPUs.
Such multi-GPU setups entail high hardware costs and considerable maintenance complexity, making them impractical for most users. This limitation is particularly critical in personal or small-scale professional settings, where 3DGS is often applied to reconstruct scenes from user-provided images, such as VR hobbyists modeling personal spaces~\cite{scaniverse, polycam, sketchfab}, interior designers designing 3D virtual rooms~\cite{arkio, worldexplorer}, and real estate professionals supporting 3D virtual tours~\cite{realhorizon}. Thus, enabling high-quality 3DGS training on a single consumer-grade GPU is essential for accessible deployment.

We present \textbf{\name}, a fast, memory-efficient, and scalable 3D Gaussian Splatting training system built upon host (CPU) offloading. Our key observation is that in each training iteration, only a small subset of Gaussian parameters participates in forward and backward passes. Leveraging this property, \name stores all Gaussian parameters and optimizer states in host memory, transferring only the necessary subset to the GPU on demand. While this approach dramatically reduces GPU memory usage,
it forces computationally intensive frustum culling and memory intensive optimizer updates onto the CPU, leading to significant slowdowns due to limited compute power and memory bandwidth of CPU. To address these challenges, \name incorporates three system-level optimizations:

\begin{itemize}
    \item \textit{Selective Offloading}: Only geometric attributes of parameters are kept on GPU for fast frustum culling, while the rest are offloaded to host memory.
    \item \textit{Parameter Forwarding}: 
    By pre-updating only necessary parameters, this optimization breaks the dependency between CPU optimizer updates and GPU forward \& backward passes, enabling pipelining.
    \item \textit{Deferred Optimizer Update}: By deferring updates for Gaussians with zero gradients, the amount of memory accesses is substantially reduced while achieving identical training results.
\end{itemize}

Through extensive evaluations across various datasets and platforms, we demonstrate that \name can train much larger scenes on consumer-grade GPUs while maintaining training speeds comparable to GPU without host offloading. For example, \name can scale the number of Gaussians from 4 million to 18 million on an RTX 4070 Mobile GPU, yielding a 35.3\% improvement in LPIPS for the Rubble scene (Figure~\ref{fig:thumbnail}).

Our contributions are summarized as follows.
\begin{itemize}

\item We empirically observe a sparse workload characteristic: during each training iteration, only a small subset of Gaussian parameters is involved in the forward and backward passes.

\item We analyze GPU memory bottlenecks in 3DGS training and identify host-offloading opportunities based on this sparse workload characteristics.

\item We propose \name, a fast, memory efficient, and scalable training system for 3DGS. To the best of our knowledge, \name is the first host offloading based training system for 3DGS.

\item We implement \name on top of gsplat~\cite{gsplat} library and comprehensively evaluate the performance on various datasets and GPU platforms. \name demonstrates substantial GPU memory savings and comparable training speed with GPU without host offloading, unlocking large-scale 3DGS training.
\end{itemize}





\section{Background}
\label{sec:background}

\subsection{Novel View Synthesis}
Novel view synthesis generates photorealistic 3D scene images from previously unseen viewpoints using a set of 2D images captured from multiple viewpoints.
It has broad applications in diverse fields such as virtual reality (VR)~\cite{vrgs, vrsplatting, fovnerf}, augmented reality (AR)~\cite{splatloc, taoavatar}, and digital twins~\cite{articulatedgs}.
Traditional explicit 3D reconstruction methods using meshes, voxels, or point clouds often struggle with complex visual phenomena and suffer quality degradation from incomplete or inaccurate reconstructions. Differentiable rendering methods have recently emerged, offering substantial improvements in reconstruction fidelity and rendering quality. The following sections describe them.

\subsection{Neural Radiance Fields (NeRF)}
Neural Radiance Fields (NeRF)~\cite{nerf} has revolutionized the field of novel view synthesis by adopting an implicit neural representation of 3D scenes as continuous volumetric functions, overcoming limitations of explicit 3D methods. NeRF represents each 3D point with 5D coordinates $(x, y, z, \theta, \phi)$, where $(x, y, z)$ denotes the 3D position and $(\theta, \phi)$ represents the viewing direction. Multi-Layer Perceptrons (MLPs) map these coordinates to a volume density $\sigma$ and a view-dependent color $c$, simultaneously modeling both the geometric structure and appearance of the scene. Rendering involves casting rays from the camera into the scene, taking multiple sample points along each ray, predicting the color $c_i$ and density $\sigma$ for each point with the MLP, and accumulating them to compute the final pixel color via a classical volume rendering. 
Training optimizes MLP-based 3D scene representations by minimizing the difference between the rendered images and the corresponding ground-truth images.

NeRF and its variants~\cite{mega-nerf, instantngp, tensorf, fastnerf} has achieved a significant breakthrough in novel view synthesis, demonstrating superior quality. However, its dependence on MLP leads to high computational cost in both training and rendering, limiting its deployment in latency-sensitive applications.

\subsection{3D Gaussian Splatting}
3D Gaussian Splatting~\cite{3dgs} is state-of-the-art differentiable rendering method representing 3D scenes with trainable 3D Gaussian primitives. Each 3D Gaussian has 59 parameters: center position $mean \in \mathbb{R}^3$, $scale \in \mathbb{R}^3$ controlling its spatial extent, rotation represented by $quaternion \in \mathbb{R}^4$, $opacity$ that determines transparency, and $spherical$ $harmonics$ (SH) coefficients which encodes view-dependent color. The SH models how a point's color changes with viewing direction; A common degree of $L = 3$ yields 16 coefficients per color channel and 48 parameters in total for RGB.

3DGS renders images by projecting 3D Gaussians onto a 2D plane and accumulating colors in depth order. The training process minimizes the difference between rendered images and ground-truth images via backpropagation (refer to Section~\ref{sec:pipeline}).
Unlike NeRF, 3DGS uses an explicit representation, eliminating the need for MLP computation during rendering and training, leading to faster speeds and higher visual quality. However, this explicit representation significantly increases the number of required parameters, leading to a much higher memory footprint for both rendering and training. The number of Gaussians directly determines the parameter count. More Gaussians are necessary for higher rendering quality, increasing memory pressure.

\begin{figure*}[t]
  \centering
  \includegraphics[page=5, width=\textwidth]{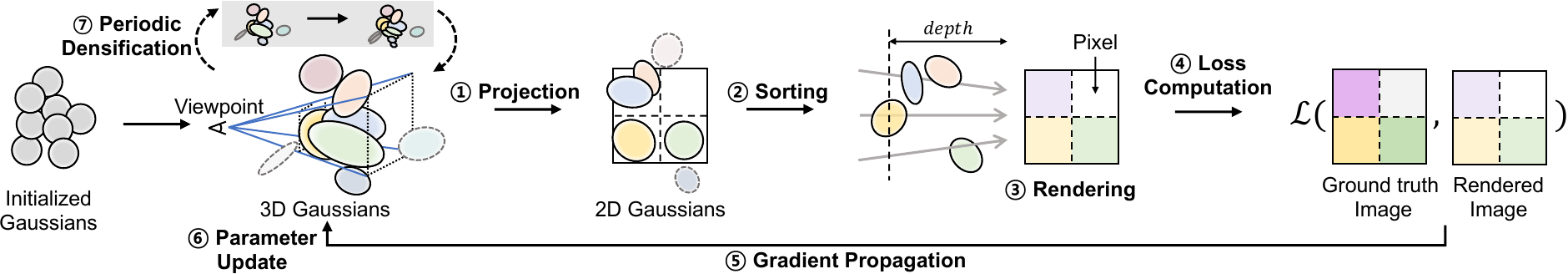}
  \caption{Overview of 3D Gaussian Splatting Training Pipeline.}
  \label{fig:background}
\end{figure*}

\subsection{Training Pipeline of 3D Gaussian Splatting}
\label{sec:pipeline}
The 3D Gaussian Splatting pipeline, illustrated in Figure~\ref{fig:background}, begins with 3D Gaussians that are initialized based on a 3D point cloud obtained from Structure-from-Motion (SfM)~\cite{sfm}.

Training iteratively performs following seven key steps. \circled{1} 3D Gaussians (ellipsoids) are projected onto the image plane, producing 2D Gaussians (ellipses). Gaussians outside the near and far planes of the viewing frustum are excluded from projection (\textit{frustum culling}).
Projection of 3D Gaussians consists of two steps. First, the geometric parameters of each 3D Gaussian (i.e., 3D $mean$, $scale$, $quaternion$) are transformed into its 2D counterparts (2D $mean$ and covariance matrix). Second, the RGB color of each 2D Gaussian is computed from the spherical harmonics coefficients and the current view direction. After the projection, \textit{frustum culling} is performed again, excluding 2D Gaussians outside the image boundaries from subsequent processing. \circled{2} The resulting 2D Gaussians are sorted by depth to ensure correct occlusion ordering. \circled{3} The color of depth sorted 2D Gaussians are blended to produce the final rendered image by using the same classical volume rendering equation as NeRF. \circled{4} The difference between the rendered image and the corresponding ground-truth image is computed to produce the loss value. \circled{5} The gradients of this loss are backpropagated. \circled{6} The backpropagated gradients are used to update the 3D Gaussian parameters.
\circled{7} Periodically (e.g., every 100 iterations), 3DGS performs densification, adaptively controlling Gaussian density to improve scene representation quality. Gaussians with large accumulated gradients are split or cloned to capture fine details, while insignificant, low-opacity Gaussians are pruned.
This step stops after a predefined iteration threshold.

\section{Motivation}

\subsection{Scaling Challenges in 3D Gaussian Splatting}
3D Gaussian Splatting (3DGS) demands significantly more memory than NeRF-based methods due to its explicit scene representation. 

\begin{figure}
     \centering
     \begin{subfigure}[b]{0.24\textwidth}
        \centering
        \noindent\includegraphics[width=\textwidth]{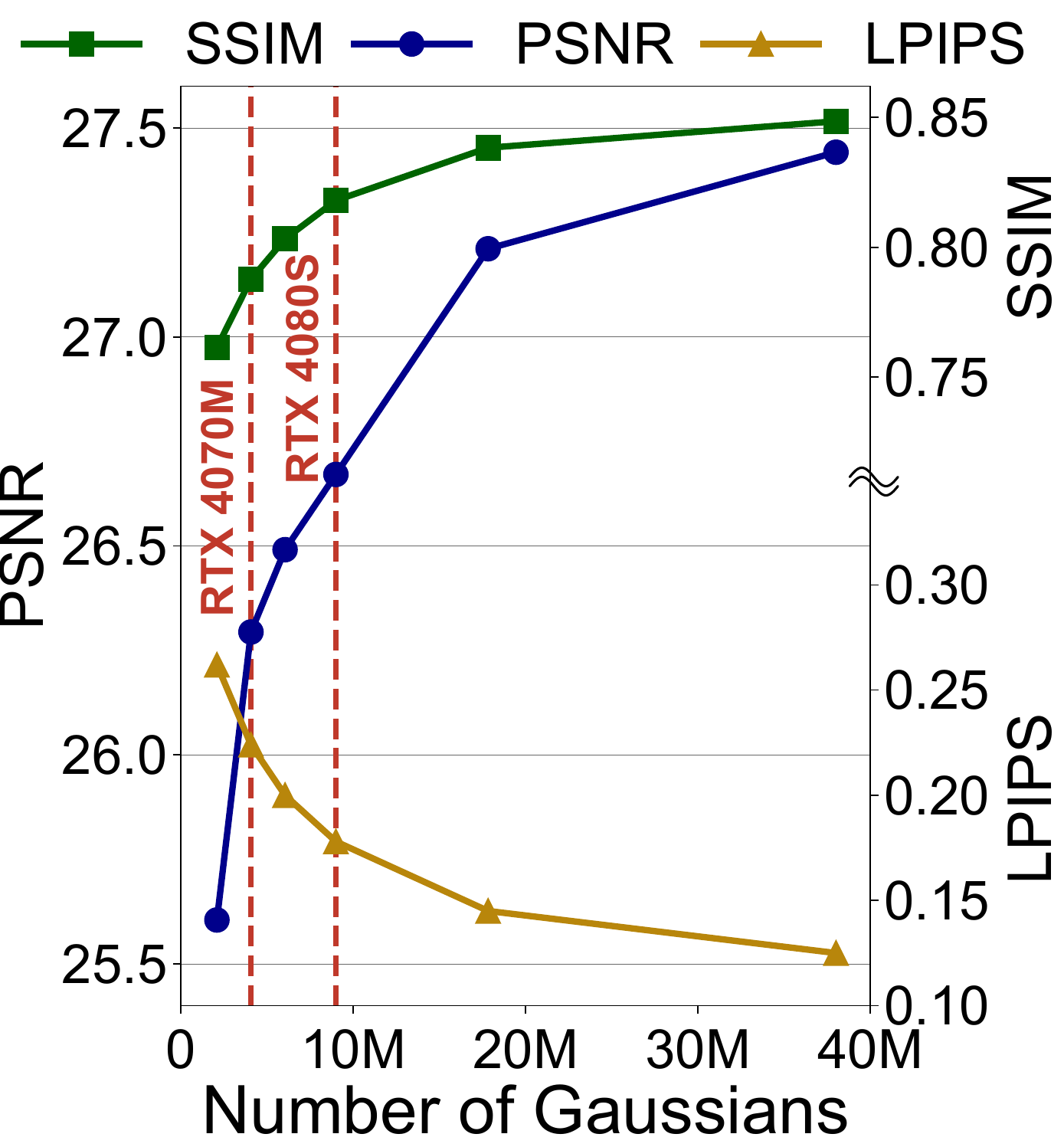} 
        \caption{Effect of the Number of Gaussians on Rendering Quality}
        \label{fig:scaling_motiv}
     \end{subfigure}
     \hfill
     \begin{subfigure}[b]{0.21\textwidth}
        \centering
        \noindent\includegraphics[width=\textwidth]{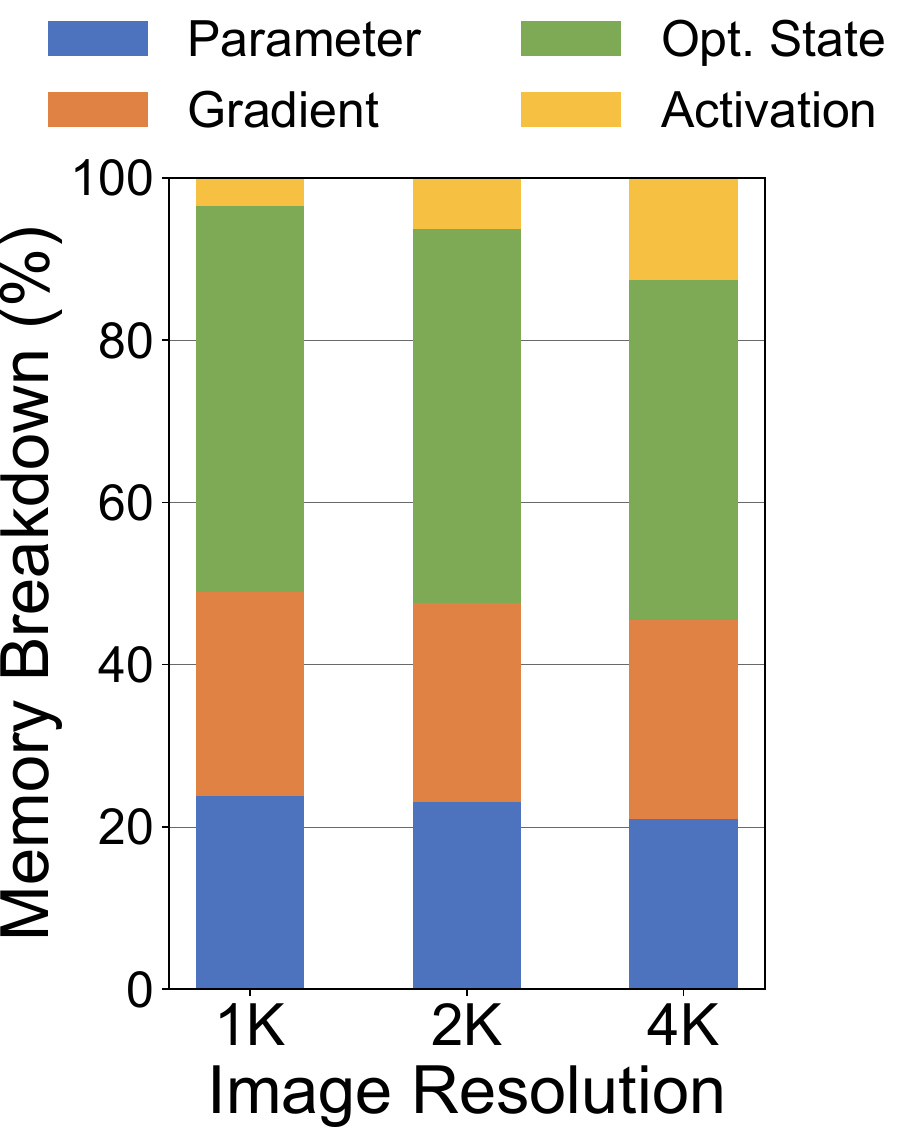}
        \caption{Breakdown of GPU Memory Usage}
        \label{fig:breakdown_memory}
     \end{subfigure}
     \caption{Scaling Challenges and Memory Bottleneck Analysis on 3D Gaussian Splatting Training.}
     \vspace{-0.1in}
\end{figure}

This memory pressure intensifies during training, consuming over four times the memory of the Gaussian parameters due to the need to store gradients, two optimizer states per parameter (momentum and variance in case of Adam optimizer), and additional activation memory. 
As demonstrated in Figure~\ref{fig:scaling_motiv}, increasing the number of Gaussians improves rendering quality, but the GPU memory limit restricts 3DGS scalability. A single RTX 4080 Super GPU can train about 9 million Gaussians, limiting the PSNR to 26.67 on Rubble scene. This limitation is critical given that 3D Gaussian Splatting is frequently used to train user-captured personalized 3D scenes, which often relies on consumer-grade GPUs with limited memory capacity.

\subsection{Memory Bottleneck Analysis in Training}
Figure~\ref{fig:breakdown_memory} shows a detailed breakdown of GPU memory usage with varying image resolutions measured on Building~\cite{mega-nerf} scene. We observe that Gaussian parameters, gradients, and optimizer states account for around 90\% of the total memory usage, while activations, used during forward and backward propagation, only comprise around 10\%. This trend becomes even more pronounced when lower image resolutions are used because activation size scales with the number of rendered pixels. Considering that 1K to 4K resolutions are commonly used in 3DGS~\cite{citygaussian, citygaussianv2, hug, mega-nerf, grendel}, reducing GPU memory usage requires targeting the Gaussian-related components rather than activations.

\subsection{Opportunities of Host Offloading}\label{sec:Opp_of_host_offloading}
\begin{figure}
  \centering
  \includegraphics[width=\columnwidth]{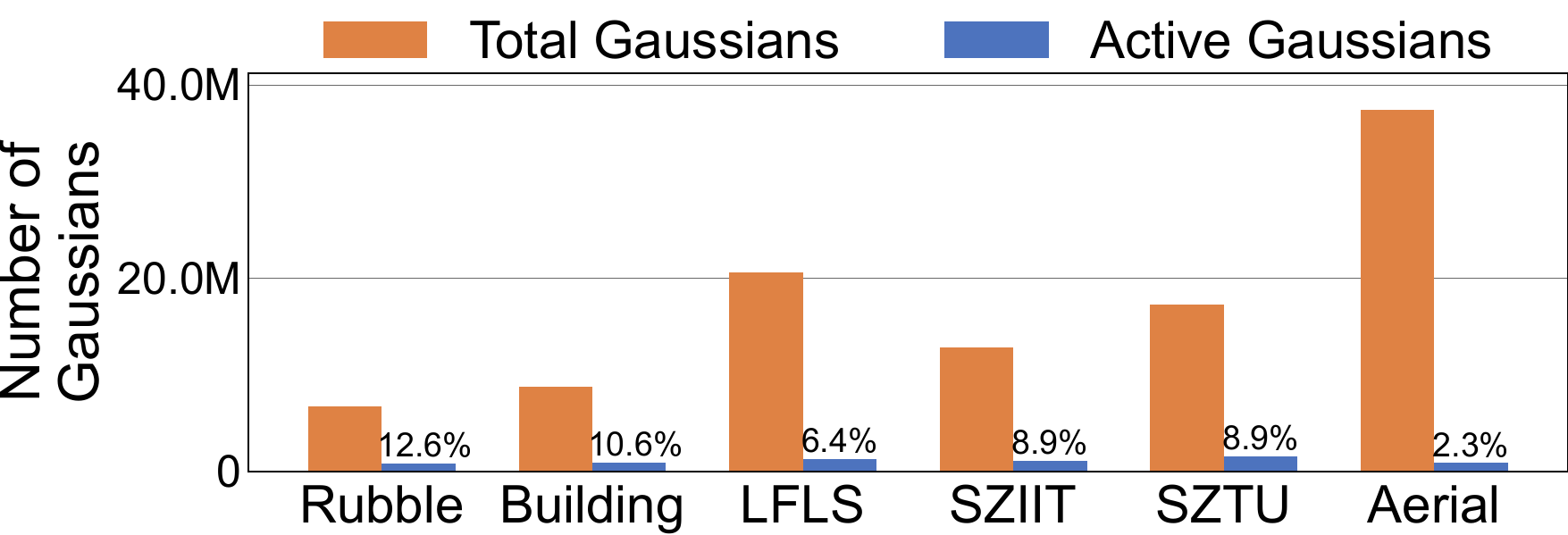}
  \caption{Average Number of Gaussians Used while Training Compared to Total Gaussians.}
  \label{fig:opportunity}
  \vspace{-0.1in}
\end{figure}

A unique characteristic of 3D Gaussian Splatting training pipeline is that only Gaussians within the viewing frustum are used for rendering (forward propagation), loss computation, and backward propagation. Our profiling results in Figure~\ref{fig:opportunity} show that each training iteration utilizes only 8.28\% of total Gaussians on average in large-scale scenes. Most training stages operate on a small subset of Gaussians, except for frustum culling, which needs access to all Gaussians, and optimizer updates, which update all parameters and optimizer states. This insight suggests offloading all Gaussian parameters and optimizer states to host (CPU) memory, transferring only necessary data to GPU memory on demand to significantly save GPU memory.

\subsection{Challenges in Host Offloading}
\label{sec:challenges}
While conceptually simple, offloading Gaussian parameters and optimizer states to host memory introduces several significant challenges as described below.

\subsubsection*{\textit{Challenge 1: Frustum culling is slow on the CPU}} 
Identifying Gaussians within the viewing frustum requires processing the entire set of Gaussians. Accurate frustum culling requires projecting each 3D Gaussian onto the 2D image plane to determine whether it lies within the image boundaries. Performing this compute-intensive operation on the CPU, with its significantly lower FLOPS compared to a GPU, introduces substantial overhead.

\subsubsection*{\textit{Challenge 2: Slow optimizer updates on CPU due to low CPU memory bandwidth and inefficient nature of Adam optimizer}}

Adam~\cite{adam}, which is the most widely used optimizer in 3DGS~\cite{3dgs, gsplat, grendel}, updates all parameters and optimizer states including those with zero gradients since its momentum terms remain nonzero even when the gradients are zero (refer to Equation~\ref{eq:adam}). 
Thus, all parameters and optimizer states must be updated by CPU, regardless of whether the corresponding Gaussians were involved in forward/backward propagation since GPU memory cannot hold them all. 
Given that optimizer updates are memory-bound and CPU memory bandwidth is typically much lower than GPU memory, this leads to considerable training slowdown.

\subsubsection*{\textit{Challenge 3: Peak memory usage is bound by the most demanding training image}} 
Although only the Gaussians within the viewing frustum of each training image are fetched on demand, the peak memory usage is determined by the image that requires the largest number of Gaussians.
Even if most training images activate a small subset of Gaussians, a single image with an exceptionally large coverage (i.e., image seen from a far viewpoint) can dominate the peak memory requirement, limiting the effectiveness of host offloading.

\section{\name Design}
This section introduces \name, our novel system designed to overcome the three challenges in the host offloading. \name leverages strategic host offloading combined with several system-level optimizations to enable efficient and scalable 3D Gaussian Splatting training on commodity GPUs.

\begin{figure}
  \centering
  \includegraphics[page=1, width=\columnwidth]{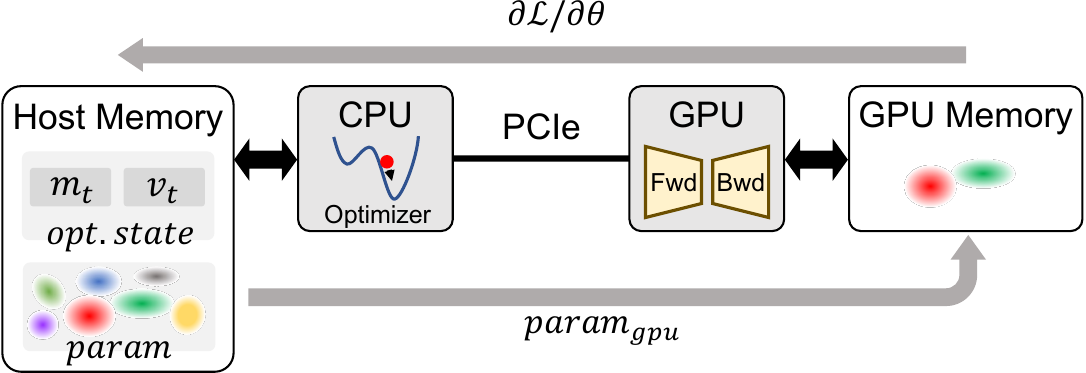}
  \caption{Baseline \name with Host Offloading.}
  \label{fig:overview}
  \vspace{-0.05in}
\end{figure}

\subsection{Baseline \name with Host Offloading}
To the best of our knowledge, no prior work has explored host (CPU) offloading to reduce GPU memory usage in 3D Gaussian Splatting training. Thus, we first implement a baseline training system that offloads Gaussians to host memory without specific optimizations proposed in Section~\ref{sec:optimization}. Figure~\ref{fig:overview} illustrates the system. All Gaussian parameters and optimizer states reside in host memory.
Only the necessary Gaussians are transferred to GPU memory
via a PCIe interconnect for forward and backward passes, with gradients then sent back to the CPU for optimizer updates.

\subsubsection*{Training Process:}
\begin{figure}[t]
  \centering
  \includegraphics[page=2, width=\columnwidth]{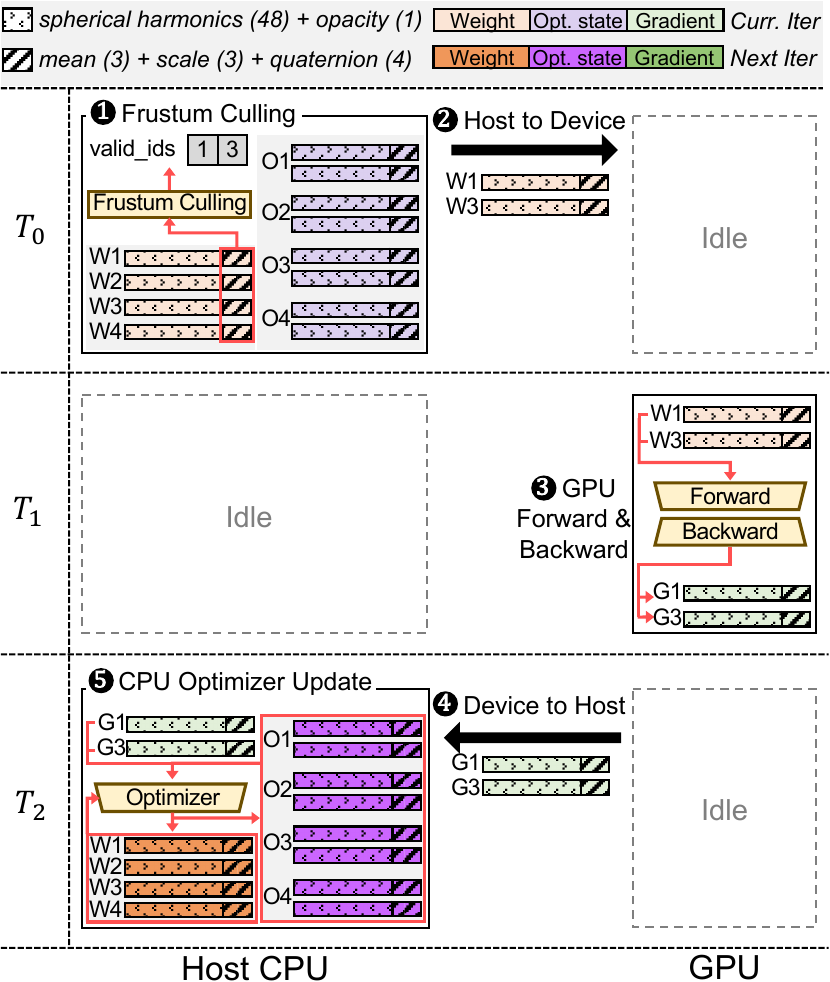}
  \caption{Training Iteration in Baseline \name.}
  \label{fig:baseline}
\end{figure}

Figure~\ref{fig:baseline} illustrates the training iteration of the baseline \name, and Figure~\ref{fig:pipeline}b shows the corresponding execution timeline. The memory state at each timestamp (i.e., $T_{0}$, $T_{1}$, $T_{2}$) in Figure~\ref{fig:baseline} is a snapshot of the system at the corresponding point in time shown in Figure~\ref{fig:pipeline}b.
\circled{1} Training begins with CPU-based frustum culling identifying Gaussians within the training image's viewing frustum (Gaussian \#1 and \#3). This step relies solely on spatial relationships, thus only geometric parameters, i.e., mean, scale, quaternion, are used (refer to the hatched area). \circled{2} The IDs of the selected Gaussians are stored in \texttt{valid\_ids}, and the corresponding parameters (i.e., W1, W3) are transferred to GPU memory via PCIe. ($t=T_{0}$). \circled{3} Forward and backward passes are performed on the GPU ($t=T_{1}$). \circled{4} The gradients (i.e., G1, G3) are transferred back to the CPU. \circled{5} Adam optimizer updates all Gaussian parameters and states on the CPU ($t=T_{2}$). Note that Adam optimizer also updates Gaussian parameters that do not receive gradients (i.e., W2, W4) because their corresponding optimizer states (i.e., O2, O4) can remain nonzero. As shown in Figure~\ref{fig:baseline}, updated weights and optimizer states are highlighted for clarity.

\subsubsection*{Performance Challenges:}
\begin{figure}
  \centering
  \includegraphics[width=\columnwidth]{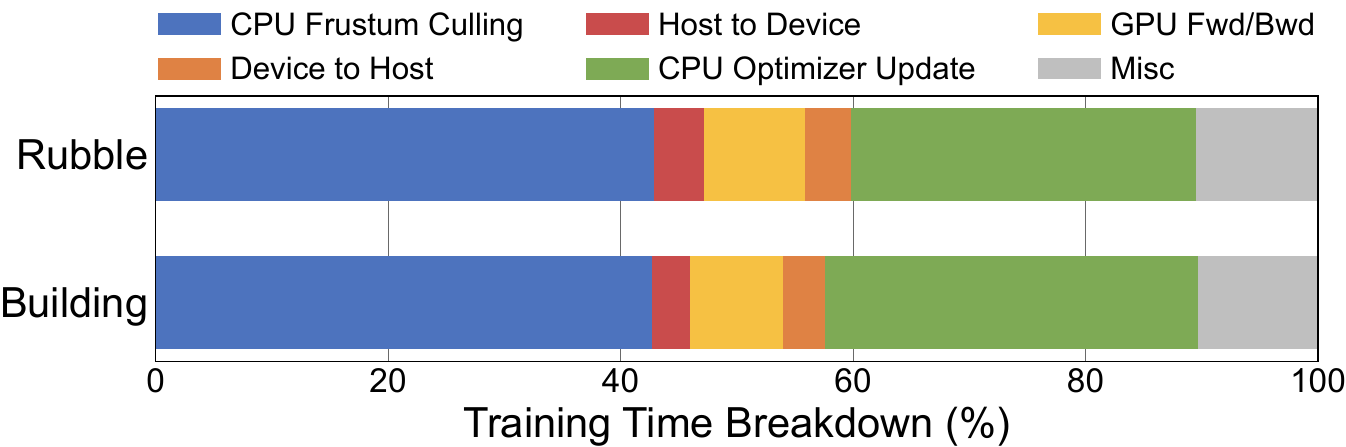}
  \caption{Training Time Breakdown of Baseline \name Measured on RTX 4070 Mobile GPU.}
  \label{fig:breakdown}
\end{figure}

Despite reducing GPU memory, this baseline system incurs significant training time overhead, making around 4$\times$ slower than GPU-only training. Figure~\ref{fig:breakdown} presents its training time breakdown, measured on a laptop with an RTX 4070 Mobile GPU. As discussed in Section~\ref{sec:challenges}, the primary bottlenecks are frustum culling and optimizer updates, both executed on the host CPU.
\begin{itemize}
    \item \textit{Slow Frustum Culling on CPU} (\circled{1}): The CPU has significantly lower compute capability (52$\times$ less peak FLOPS on ASUS TUF Gaming F17 laptop) compared to the GPU, making the compute-intensive frustum culling a major bottleneck when performed on CPU.
    \item \textit{Slow Optimizer Updates on CPU} (\circled{5}): The CPU's memory bandwidth is 3$\times$ slower than the GPU's, which turns memory-intensive optimizer updates into a major bottleneck when executed on the CPU.
    \item \textit{GPU Idle Time Due to Dependency} (\circled{3}, \circled{5}): A dependency exists between GPU-based forward/backward propagation and CPU-based optimizer updates, causing the GPU to remain idle for a significant amount of time during CPU execution. 
\end{itemize}

\subsection{\name Optimizations}\label{sec:optimization}
To address these bottlenecks, we will explore various system-level optimization opportunities in the subsequent sections. These optimizations leverage the unique characteristics of the 3D Gaussian Splatting training pipeline (Section~\ref{sec:Opp_of_host_offloading}).

\subsubsection{Selective Offloading}
To mitigate the CPU-based frustum culling bottleneck, we propose \textit{selective offloading}, moving this operation to the GPU. Since only the position and the size of Gaussians (i.e., mean, scale, quaternion) are needed to determine visibility within the viewing frustum, only the geometric attributes of all parameters are kept on the GPU for fast frustum culling. The geometric attributes comprise only 10 out of 59 Gaussian parameters, resulting in a modest 17\% GPU memory overhead. This is a worthwhile trade-off for significantly faster GPU-based frustum culling and reduced training time. Also, the non-geometric attributes (83\%) are offloaded to host memory, still achieving considerable memory saving.

\begin{figure}[t]
  \centering
  \includegraphics[page=3, width=\columnwidth]{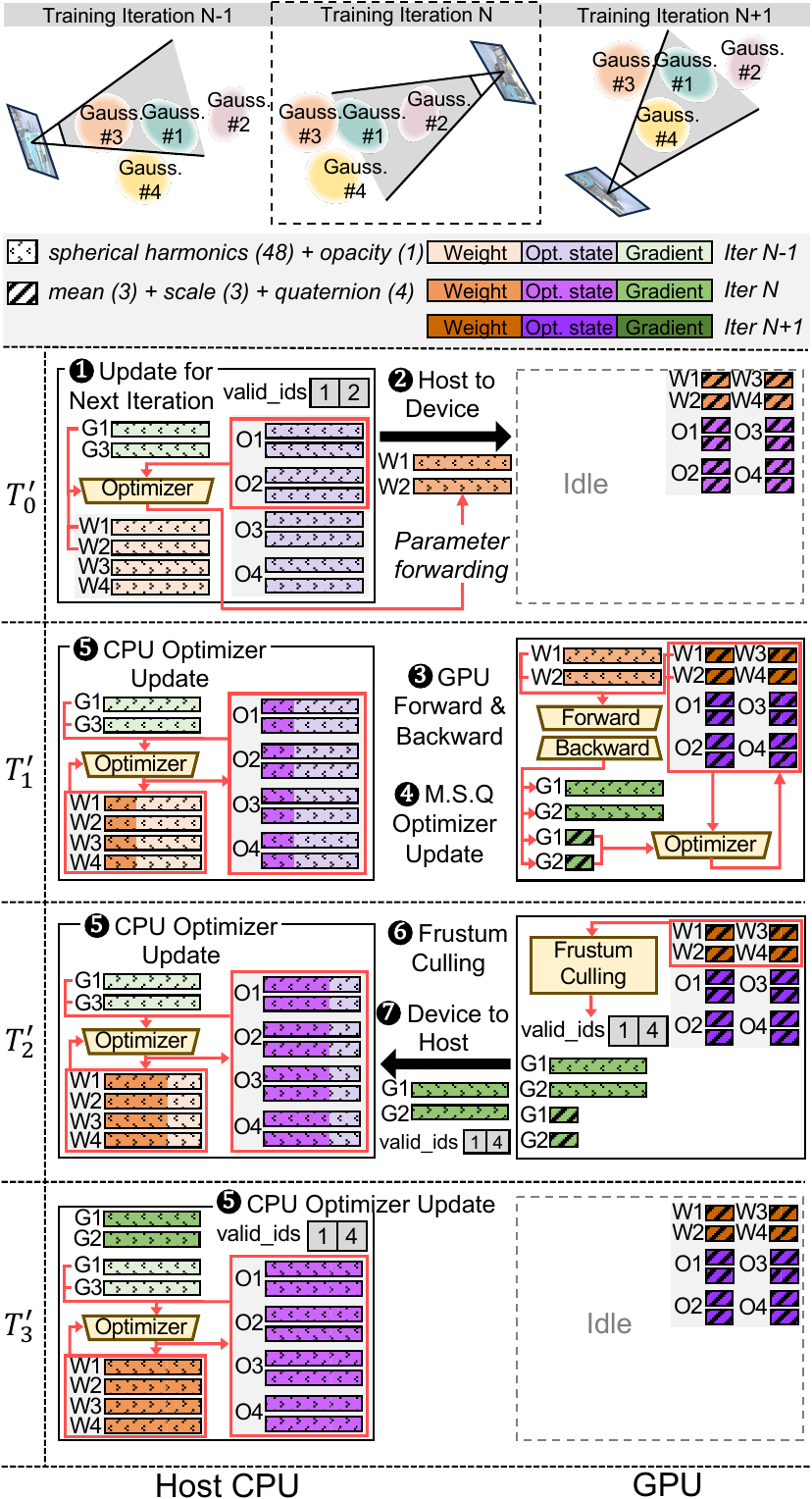}
  \caption{$N^{th}$ Training Iteration in \name with Selective Offloading and Parameter Forwarding.}
  \label{fig:ours}
  \vspace{-0.1in}
\end{figure}

\begin{figure}[t]
  \centering
  \includegraphics[page=4, width=\columnwidth]{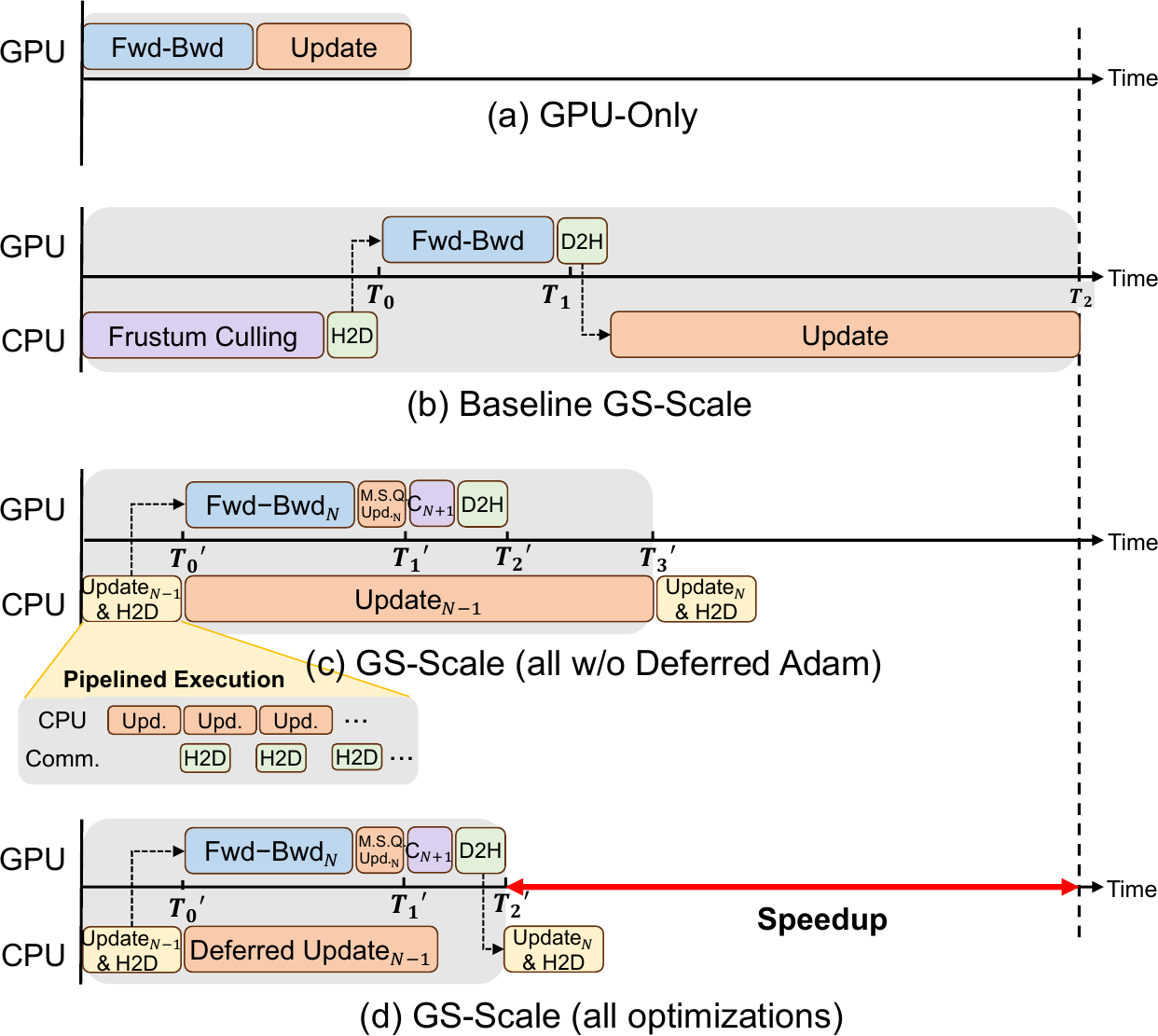}
  \caption{Execution Timeline of GPU-Only, Baseline \name, and \name with optimizations. H2D/D2H denote Host to Device transfers, M.S.Q. is mean/scale/quaternion.}
  \label{fig:pipeline}
  \vspace{-0.1in}
\end{figure}

\subsubsection{Breaking Data Dependency via Parameter Forwarding}
Optimizer updates become the primary bottleneck after selective offloading.  To mitigate this overhead, we introduce a pipelined training scheme, enabling concurrent execution of forward/backward passes on GPU and CPU optimizer updates. Typically, such pipelining is not feasible due to the data dependency since updated parameters are needed for the the next training iteration’s forward pass. 

We identify a unique opportunity offered by 3D Gaussian Splatting training workloads. Each training iteration requires only a small subset of parameters corresponding to Gaussians within the viewing frustum of the next training image. We exploit this with \textit{parameter forwarding}, which performs early updates of only the parameters needed for the next iteration on the CPU and forwards the updated parameters to the GPU. Remaining parameters are updated asynchronously on the CPU in a lazy manner, enabling pipelining with forward/backward passes on GPU. 

Figure~\ref{fig:ours} demonstrates the detailed working example of \name with both \textit{selective offloading} and \textit{parameter forwarding}. 
The memory state snapshot of Figure~\ref{fig:ours} ($T_{0}'$, $T_{1}'$, $T_{2}'$, and $T_{3}'$) corresponds to timestamps on the execution timeline of Figure~\ref{fig:pipeline}c.
We assume that the forward and backward propagation for visible Gaussians \#1 and \#3 are complete as a part of the $(N-1)^{th}$ training iteration. Their gradients (G1 and G3) have been generated and stored in host memory at $T_{0}'$. 
The frustum culling operation for the $N^{th}$ training iteration is also complete, identifying visible Gaussians \#1 and \#2 and assigning 1 and 2 to \texttt{valid\_ids}. These \texttt{valid\_ids} are used for the forward and backward propagation of the $N^{th}$ training iteration.

\circled{1} \textit{Parameter forwarding} updates only the parameters required for the current iteration (W1 and W2) by using the corresponding gradient (G1) from the previous iteration; G2 is zero at this point.
\circled{2} The updated parameters are transferred to the GPU via PCIe; CPU-side copies of the parameters and optimizer states remain unchanged ($t=T_{0}'$). 
To mitigate the transfer overhead, parameters are partitioned into 32MB chunks, enabling pipelined execution between CPU-side optimizer updates and host-to-device transfers. The execution timeline in Figure~\ref{fig:pipeline}c illustrates this scheme. 
\circled{3} Once transferred, the GPU executes forward/backward passes using these parameters and geometric parameters (mean, scale, quaternion of Gaussian \#1 and \#2) already on the GPU via \textit{selective offloading}. 
\circled{4} Since geometric parameters and the corresponding optimizer states always reside in the GPU, they are immediately updated after the forward/backward pass ($t=T_{1}'$). \circled{5} Concurrently ($t=T_{1}'$), remaining parameters and their optimizer states, including non forwarded ones to the GPU, are lazily updated on the CPU, minimizing GPU idle time. Note that this optimizer process is a part of the $(N-1)^{th}$ training iteration.
\circled{6} Once the GPU-side geometric parameters are updated, frustum culling is performed using the updated parameters and the next training image, identifying the visible Gaussians for the $(N+1)^{th}$ iteration. 
\circled{7} Finally, gradients of non-geometric parameters are transferred back to CPU and held until the CPU completes its optimizer updates ($t=T_{3}'$).

\subsection{Deferred Optimizer Update}
While parameter forwarding enables pipelining between CPU and GPU, a slow CPU-based optimizer can still dominate overall execution, as shown in Figure~\ref{fig:pipeline}c. This is mostly due to low CPU memory bandwidth and Adam optimizer's inefficiency (updating all states and parameters, even those with zero gradients). To further accelerate the CPU-based Adam optimizer without algorithmic changes, we propose \textit{deferred optimizer update}. Although we use Adam as an example, \textit{deferred optimizer update} can be extended to most momentum-based optimizers, such as SGD (stochastic gradient descent) with momentum and AdamW~\cite{adamw}.

\subsubsection{Optimization Opportunities} 
\label{sec:opportunity}
We can defer updates for parameters and optimizer states with zero gradients because their values can be precisely reconstructed by tracking deferred iterations. This is due to momentum based optimizer's deterministic behavior when gradients are zero, as shown in Adam's example (Equation~\ref{eq:adam}). If gradient $g_t$ is zero, momentum and variance $m_t$ and $v_t$ are simply scaled by ${\beta}_1$ and ${\beta}_2$, respectively.

\begin{equation}
\small
\label{eq:adam}
\begin{aligned}
m_t &= \beta_1 m_{t-1} + (1 - \beta_1) g_t, \quad \hat{m}_t = \frac{m_t}{1 - \beta_1^t} \\
v_t &= \beta_2 v_{t-1} + (1 - \beta_2) g_t^2, \quad \hat{v}_t = \frac{v_t}{1 - \beta_2^t}\\
w_{t+1} &= w_t - \frac{\eta}{\sqrt{\hat{v}_t} + \epsilon} \hat{m}_t
\end{aligned}
\end{equation}

This property enables us to restore current optimizer states ($m_{t}$ and $v_{t}$) from deferred optimizer states ($m_{t-d-1}$ and $v_{t-d-1}$, $t>d$) and the defer count $d$. If the gradient remains zero for $d$ consecutive iterations and becomes non-zero at iteration $t$, momentum and variance can be reconstructed by simply multiplying scaling factors as follows:

\begin{equation}
\small
\label{eq:deferred_adam_1}
\begin{aligned}
m_t &= \underbrace{\beta_1^{d+1}}_{\text{\texttt{m\_scale}}} m_{t-d-1} + (1 - \beta_1) g_t \\
v_t &= \underbrace{\beta_2^{d+1}}_{\text{\texttt{v\_scale}}} v_{t-d-1} + (1 - \beta_2) g_t^2 \\
\end{aligned}
\end{equation}

Parameter $w_{t}$ can also be restored from deferred parameter $w_{t-d}$, momentum $m_{t-d-1}$, and variance $v_{t-d-1}$ by repeatedly applying the weight update $d$ times:

\begin{equation}
\small
\label{eq:deferred_adam_2}
\begin{aligned}
w_t &= w_{t-d}
 - \sum_{l=0}^{d-1} 
   \frac{\eta}{\sqrt{\hat{v}_{t-d+l}} + \epsilon} \, \hat{m}_{t-d+l} \\
&= w_{t-d}
 - \sum_{l=0}^{d-1} 
   \frac{\eta}{
      \sqrt{
        \frac{\beta_2^{l+1} v_{t-d-1}}{1 - \beta_2^{t-d+l}}
      } + \epsilon
   }
   \cdot
   \frac{\beta_1^{l+1} m_{t-d-1}}{1 - \beta_1^{t-d+l}} \\
&\approx w_{t-d}
 - \frac{m_{t-d-1}}{\sqrt{v_{t-d-1}} + \epsilon}
   \underbrace{
   \sum_{l=0}^{d-1}
   \frac{\eta}{
      \sqrt{\frac{\beta_2^{l+1}}{1 - \beta_2^{t-d+l}}}
   }
   \cdot
   \frac{\beta_1^{l+1}}{1 - \beta_1^{t-d+l}}
   }_{\text{\texttt{w\_scale}}}\\
\end{aligned}
\end{equation}

Assuming that $\epsilon$ is small, we can factor out $m_{t-d-1}$ and $v_{t-d-1}$, making the remaining expression (i.e., \texttt{w\_scale}) a precomputable constant, which simplifies weight restoration. Note that $\epsilon$ is typically a very small constant introduced to prevent divide-by-zero errors and this approximation has negligible effect on training, which we substantiate in Section~\ref{sec:correctness}. 
Finally, the restored $w_{t}$, $m_{t}$, and $v_{t}$ are used to produce final weight $w_{t+1}$ with the original Adam rule.

\subsubsection{Implementation}
We propose the \textit{deferred optimizer update}, which defers updates for Gaussians not involved in forward and backward propagation. Instead of immediate updates, a 4-bit counter increments for deferred updates, allowing up to 15 deferrals. Parameters and optimizer states are restored only when their corresponding gradient becomes non-zero or the counter reaches its maximum. Even with conservative estimates, this results in only 6.7\% (1/15) unnecessary updates due to counter saturation.

\begin{figure}[t]
    \centering
    \begin{minipage}[b]{1.0\hsize}
        \begin{lstlisting}[escapechar=\^,style=myStyle2,mathescape=true]
// N: Number of total Gaussians
// D: Dimension of each parameter
float param[N][D], grad[N][D], mom[N][D], var[N][D];
char counter[N]; int MAX = 15; // MAX: Max counter value
float lr, b1, b2, eps; // Hyperparameters

// n: Number of Gaussians with nonzero gradient
$\textbf{def}$ $\textbf{deferred\_update}$(vector<int> valid_ids[n], int step):
  /* Gaussian IDs that need to be restored */
  vector<int> update_ids;
  update_ids = union(valid_ids, where(counter == MAX));

  /* Precompute scaling factor */
  float param_lut[MAX], mom_lut[MAX], var_lut[MAX];
  float scale = b1 / $\textbf{sqrt}$(b2);
  param_lut[0] = 0;
  for i = 1 to MAX:
    param_lut[i] = scale*param_lut[i-1] + 
                      (lr*b1) / ($\textbf{sqrt}$(b2/(1$\minus$$\textbf{pow}$(b2, step$\minus$i))) 
                        * (1 - $\textbf{pow}$(b1, step$\minus$i)));
  for i = 0 to MAX:
    mom_lut[i] = $\textbf{pow}$(b1, i+1);
    var_lut[i] = $\textbf{pow}$(b2, i+1);
  
  /* Perform optimizer update */
  float bias_correction = $\textbf{sqrt}$(1 - $\textbf{pow}$(b2, step));
  float step_size = lr / (1 - $\textbf{pow}$(b1, step));
  for id in update_ids:
    float delay = counter[id];
    float w_scale = param_lut[delay];
    float m_scale = mom_lut[delay];
    float v_scale = var_lut[delay];
    for k = 0 to D:
      float w = param[id][k];   float g = grad[id][k];
      float m = mom[id][k];     float v = var[id][k];
      float m_new = m_scale*m + (1$\minus$b1)*g;
      float v_new = v_scale*v + (1$\minus$b2)*g*g;
      mom[id][k] = m_new;
      var[id][k] = v_new;
      w $\minus$= (w_scale * m) / ($\textbf{sqrt}$(v) + eps);
      float denom = $\textbf{sqrt}$(v) / bias_correction + eps;
      param[id][k] = w $\minus$ step_size * m_new / denom;

  /* Update counter for deferred Gaussians */
  for id = 0 to N:
    counter[id] += 1;
  for id in update_ids:
    counter[id] = 0;
}
        \end{lstlisting}
    \end{minipage}
    \caption{Pseudocode of Deferred Optimizer Update.}
    \label{fig:deferredadam}
    \vspace{-0.1in}
\end{figure}

Figure~\ref{fig:deferredadam} details the pseudocode. A set of Gaussians to be updated (\texttt{update\_ids}) is determined as the union of those with nonzero gradients (\texttt{valid\_ids}) and those whose \texttt{counter} has reached \texttt{MAX} (Line 11). Three scaling factors for parameter and optimizer state restoration are precomputed and stored in lookup tables for each deferred step $d$ (Line: 14–23), following the equations in Section~\ref{sec:opportunity}. For each Gaussian in \texttt{update\_ids}, its defer count is read (Line 29), scaling factors are retrieved (Line: 30-32), parameters and states are reconstructed (Line: 34-40), and the standard Adam update is applied (Line: 41–42). Counters for updated Gaussians are then reset, while deferred ones increment by 1 (Line: 45-48).

\textit{Deferred optimizer update} significantly reduces memory accesses, proportional to the ratio of used to total Gaussians, while incurring minimal overhead. Each counter lookup/update requires a single 8-bit memory access (\texttt{char} datatype) per Gaussian. However, a full optimizer update requires $7D * 32$-bit accesses per Gaussian ($4D$ reads and $3D$ writes for parameters, gradients, and optimizer states, where $D$ is parameter dimension, i.e., 59; refer to Line 33-42). Parameter and optimizer state restoration adds some computation but incurs no additional memory accesses (Line 40), thus having little impact on overall execution time, as optimizer updates are primarily memory-bound.

\subsubsection{Integration to \name} 
Deferred optimizer update integrates into the \name pipeline with minor adjustments to \textit{parameter forwarding}. 
Since forwarded parameters must be accurate, weight restoration is performed before forwarding the parameters. Neither CPU-stored original parameters nor counters are modified during this parameter forwarding process (\circled{1} in Figure~\ref{fig:ours}), while they are updated in the actual CPU optimizer update process (\circled{5} in Figure~\ref{fig:ours}).

\begin{table}
\caption{Specifications of GPU Platforms}
\setlength{\tabcolsep}{3.5pt}
    \centering
    \footnotesize
    \begin{tabular}{c|c|c|c|c|c|c}
        \multirow{2}{*}{\textbf{GPU}} &
        \multicolumn{2}{c|}{\textbf{GPU Memory}} &
        \multirow{2}{*}{\shortstack{\textbf{PCIe}\\\textbf{BW}}} &
        \multicolumn{2}{c|}{\textbf{Host Memory}} & 
        \multirow{2}{*}{$\boldsymbol{R_{bw}}$}\\
        \cline{2-3} \cline{5-6}
        & \textbf{Size} & \textbf{BW} & & \textbf{Size} & \textbf{BW} & \\
        \hline\hline
        \multicolumn{6}{c}{\textbf{Laptop}} \\
        \hline
        RTX 4070M  & 8 GB   & 256 GB/s   & 16 GB/s  & 32 GB   & 83.2 GB/s & 3.1 \\
        \hline
        \multicolumn{6}{c}{\textbf{Desktop}} \\
        \hline
        RTX 4080S  & 16 GB  & 736 GB/s & 32 GB/s  & 64 GB   & 89.6 GB/s & 8.2 \\
        \hline
        \multicolumn{6}{c}{\textbf{Server}} \\
        \hline
        H100 80GB   & 80 GB  & 2.04 TB/s  & 64 GB/s  & 1 TB    & 614.4 GB/s & 3.3 \\
        \hline
    \end{tabular}
    \label{table:platform_specs}
\end{table}

\subsection{Balance-Aware Image Splitting Training}
Even with \name's significant GPU memory savings, peak memory usage is bound by the the maximum number of Gaussians from the most demanding training image. 
To address such cases, we propose \textit{balance-aware image splitting training}. When the ratio of active to total Gaussians exceeds a predefined threshold \texttt{mem\_limit}, an image is spatially partitioned into two sub-regions, each processed separately. Each sub-region undergoes independent frustum culling, followed by separate forward and backward passes to compute individual losses and gradients. The gradients are transferred to CPU immediately after they are computed and are later aggregated on the CPU, mitigating GPU memory pressure. Optimizer update is applied once for both regions using the aggregated gradients on the CPU, ensuring mathematical equivalence to the original training pipeline. Splitting a demanding image into two can halve memory usage during forward/backward passes, preserving memory savings. While more splits are possible, two sufficed in our benchmarks.

Finding an optimal split point is critical to balance Gaussian counts, as naive equal-area splitting often leads to imbalance due to varying Gaussian density. Our \textit{balance-aware image splitting strategy}, applied once before training using the initial 3D Gaussians, addresses this. We efficiently balance counts by starting at the image midpoint, performing frustum culling on both sides, and then iteratively adjusting the split toward the less-populated side via a 5-step binary search. This process adds only 0.08\% overhead to total training time. Despite slight changes resulting from densification, our benchmarks show an average split point ratio of 0.551:0.449, maintaining balance throughout training.

\begin{table}
\caption{Evaluated Benchmark Scenes}
\setlength{\tabcolsep}{6pt}
\centering
\footnotesize
{
    \begin{tabular}{c|c|c|c}
    \textbf{Dataset} & \textbf{Scene} & \textbf{Resolution} & \textbf{Type} \\
    \hline
    \hline
    \multirow{2}{*}{Mill-19~\cite{mega-nerf}} 
        & Rubble & \multirow{2}{*}{$1152 \times 864$} & \multirow{2}{*}{Real World \& Outdoor} \\
        & Building & &  \\ 
    \hline
    \multirow{3}{*}{GauU-Scene~\cite{gauuscene}} 
        & LFLS & \multirow{3}{*}{$1600 \times 1064$} & \multirow{3}{*}{Real World \& Outdoor} \\ 
        & SZIIT & &  \\ 
        & SZTU & &  \\ 
    \hline
    \multirow{1}{*}{MatrixCity~\cite{matrixcity}} 
        & Aerial   & $1600 \times 900$ & Synthetic \\ 
    \hline
    \end{tabular}
}
\label{table:dataset_details}
\end{table}

\section{Evaluation}
\label{sec:metho}

\subsection{Methodology}
We build \name on gsplat v1.5.0~\cite{gsplat}, a popular PyTorch~\cite{pytorch} based 3D Gaussian Splatting framework, which achieves state-of-the-art performance in terms of both training speed and memory usage. We implement pipelined CPU–GPU execution using Python’s \texttt{threading} module and \textit{deferred optimizer update} as a custom C++ PyTorch extension with OpenMP parallelization. The source code of \name is available at \url{https://github.com/SNU-ARC/GS-Scale.git}. Experiments are conducted primarily on laptop and desktop platforms, with additional evaluation on a server. We use ASUS TUF Gaming F17 laptop~\cite{laptop} with Intel Core i7-13620H CPU and RTX 4070 Mobile GPU, desktop with Intel Core i9-13900K CPU and RTX 4080 Super GPU, and server with 2$\times$Intel Xeon Gold 6530 CPU and H100 PCIe 80GB GPU. Table~\ref{table:platform_specs} shows detailed specifications. $R_{bw}$~\cite{decdec} denotes the ratio of GPU memory bandwidth to that of CPU. All platforms use CUDA 12.4 and PyTorch 2.2.0.

\begin{figure*}[t]
  \centering
  \includegraphics[width=\textwidth]{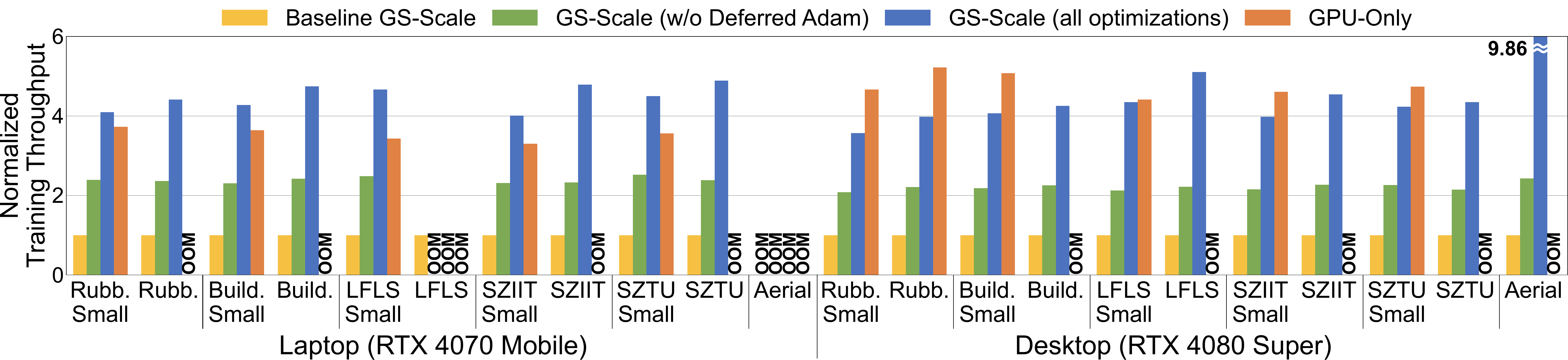}
  \caption{Training Throughput Normalized to Baseline \name.}
  \label{fig:throughput}
\end{figure*}

We evaluate \name on large-scale datasets (Table ~\ref{table:dataset_details}).  
We use 4$\times$ downsampled images for Mill-19 dataset and 1.6k resolution downsampled images for the other datasets following previous works~\cite{citygaussian, citygaussianv2, hug, mega-nerf}. Following the original 3DGS recipe, we use a batch size of 1, as larger batches offer minimal throughput gains due to limited parameter sharing. We use three standard rendering quality metrics: Peak Signal-to-Noise Ratio (PSNR), Structural Similarity Index Measure (SSIM), and Learned Perceptual Image Patch Similarity (LPIPS), where higher PSNR/SSIM and lower LPIPS indicate better quality. GPU peak memory usage is measured via PyTorch CUDA Memory Management APIs\footnote{GPU peak memory is measured based on allocated memory. Since PyTorch maintains reserved memory pools larger than the allocated memory to reduce allocation/deallocation overhead, OOM errors may occur before allocated memory reaches the GPU memory capacity.}.

We adjust densification settings (i.e., stop iteration, densification threshold, and split/clone decision threshold) to scale up or scale down Gaussian counts for each scene, following the Grendel's methodology~\cite{grendel}. A \texttt{mem\_limit} of 0.3 is used for all experiments, splitting images when active Gaussians exceed 30\% of the total.

\begin{figure}
  \centering
  \includegraphics[width=\columnwidth]{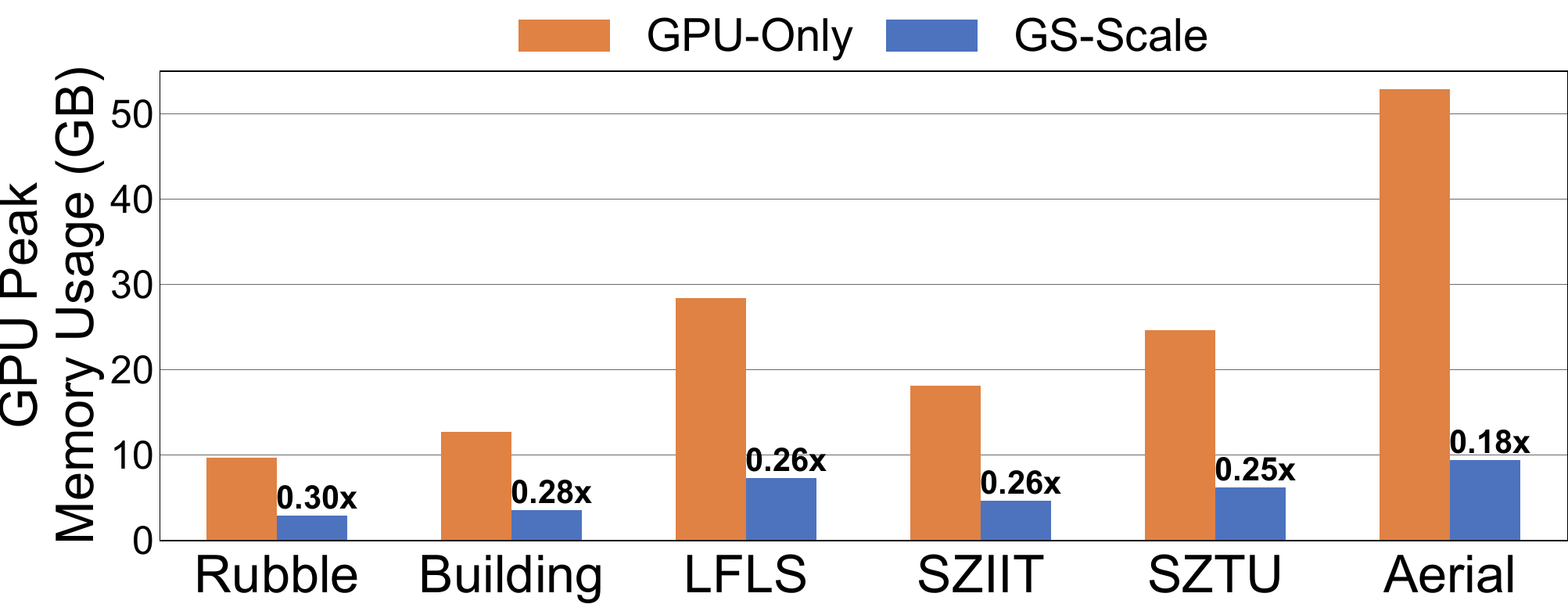}
  \caption{Peak GPU Memory Usage Savings with \name.}
  \label{fig:mem_reduction}
\end{figure}

\subsection{Memory Savings}
\label{sec:memory_savings}
We evaluate \name's memory savings against the GPU-Only system. Figure~\ref{fig:mem_reduction} demonstrates that \name achieves a substantial 3.98$\times$ geomean reduction in peak memory usage across all datasets. The memory savings correlate with the ratio of used to total Gaussians, explaining the greatest improvement in the Aerial scene. Further reduction is limited in the Aerial scene despite its low used Gaussian ratio because 17\% of parameters and optimizer states remain resident on the GPU due to \textit{selective offloading}.

\subsection{Training Throughput and Memory Efficiency}
\label{sec:throughput}
Figure~\ref{fig:throughput} evaluates training throughput across four systems: (1) baseline \name, (2) \name with all optimizations except deferred optimizer update, (3) \name with all optimizations, and (4) GPU-Only system without host offloading. Six scenes are evaluated across laptop and desktop platforms and training speed is measured in epoch time. We make smaller versions of each scene by adjusting densification settings to enable throughput comparisons.
However, we could not create a smaller version that fits into GPU memory for the Aerial scene, as its Gaussian count is already too large at initialization. Since our downsizing strategy~\cite{grendel} relies on limiting densification, scenes that trigger OOM errors before densification cannot be further downsized.

GPU-only system frequently encounters OOM errors due to limited memory, but \name's significant memory savings enable much larger-scale 3DGS training. For instance, the Aerial scene alone demands over 50GB of GPU memory without host offloading (Section~\ref{sec:memory_savings}), causing OOM on both GPU-only systems. However, \name reduces peak GPU memory usage by 5.5$\times$, allowing the Aerial scene to be trained on an RTX 4080 Super desktop. Furthermore, \name achieves comparable training throughput to GPU-Only systems, reaching geomean of 1.22$\times$ (laptop) and 0.84$\times$ (desktop) of GPU-Only performance (excluding OOM cases). 
\textit{A takeaway is that \name enables much larger scene training and consistently maintains high training throughput, even as GPU-Only systems frequently encounter OOM errors.}

\subsection{Impact of Proposed Optimizations}
Figure~\ref{fig:throughput} shows how \name's optimizations improve the training throughput over the baseline \name. We see a geomean improvement of 4.47$\times$ on laptop and 4.57$\times$ on desktop (excluding OOM cases), demonstrating \name's effectiveness.
The performance of \name depends on two key factors: (1) the GPU to CPU memory bandwidth ratio ($R_{bw}$) and (2) the average ratio of used to total Gaussians (Figure~\ref{fig:opportunity}). PCIe bandwidth also has some effect, but its impact is limited as it accounts for only a small portion of the overall training time. On platforms with lower $R_{bw}$ (like laptops), \name can perfectly pipeline CPU optimizer updates with GPU execution, even surpassing GPU-Only speeds where operations are executed sequentially. This is because lower GPU memory bandwidth slows down the memory bound backward pass (i.e., gradient accumulation) on GPU, providing enough time for CPU updates to be pipelined.
Furthermore, a lower ratio of used to total Gaussians amplifies the benefits of the deferred optimizer update, as memory access reduction scales with this ratio, explaining the notable speedups in Aerial and LFLS scenes.

\begin{figure*}
  \centering
  \includegraphics[width=0.92\textwidth]{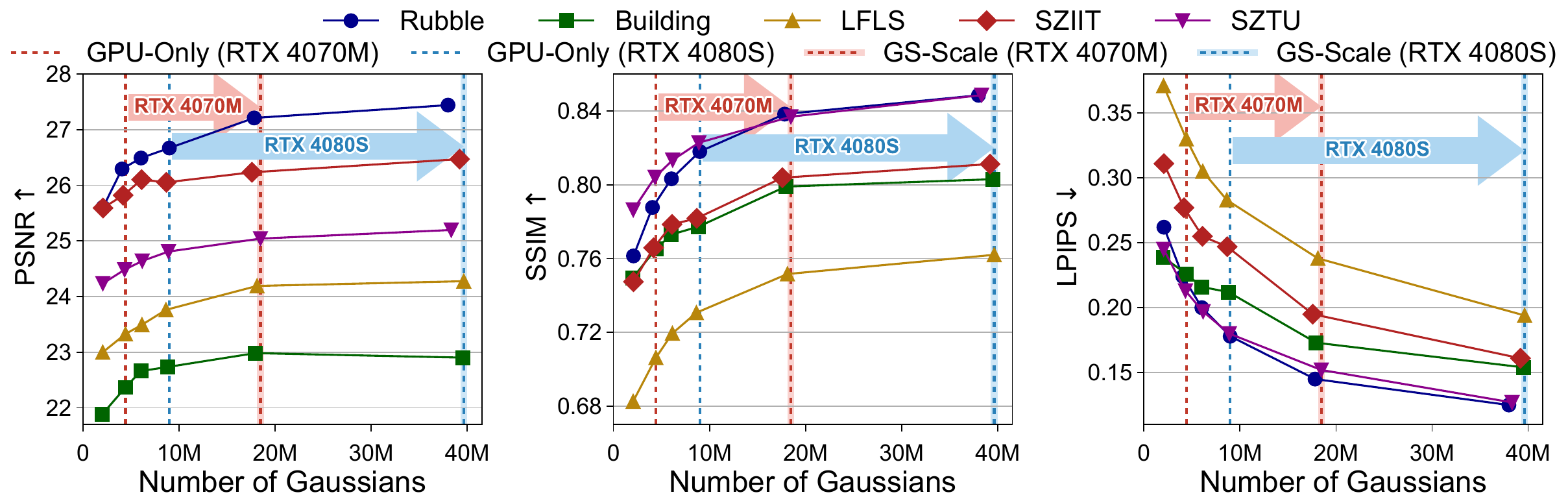}
  \caption{Evaluation of \name's Rendering Quality and Scalability Across Gaussian Scales.}
  \label{fig:scalability}
\end{figure*}

\begin{table}[t]
\caption{Impact of \name on Training Quality.}
\setlength{\tabcolsep}{10pt}
\centering
\footnotesize
{
    \begin{tabular}{c|c|c|c|c}
    \textbf{Scene} & \textbf{Method} & \textbf{PSNR\textsuperscript{$\uparrow$}} & \textbf{SSIM\textsuperscript{$\uparrow$}} & \textbf{LPIPS\textsuperscript{$\downarrow$}} \\
    \hline
    \hline
    \multirow{2}{*}{Rubble} & Original & 26.63 & 0.808 & 0.194 \\ 
    & \name & 26.62 & 0.808 & 0.194 \\ 
    \hline
    \multirow{2}{*}{Building} & Original & 22.74 & 0.777 & 0.211 \\ 
    & \name & 22.78 & 0.777 & 0.211 \\ 
    \hline
    \multirow{2}{*}{LFLS} & Original & 24.04 & 0.752 & 0.234 \\ 
    & \name & 24.08 & 0.752 & 0.233 \\ 
    \hline
    \multirow{2}{*}{SZIIT} & Original & 26.28 & 0.797 & 0.213 \\ 
    & \name & 26.29 & 0.797 & 0.213 \\ 
    \hline
    \multirow{2}{*}{SZTU} & Original & 24.90 & 0.835 & 0.155 \\ 
    & \name & 24.95 & 0.836 & 0.155 \\ 
    \hline
    \multirow{2}{*}{Aerial} & Original & 27.69 & 0.873 & 0.127 \\ 
    & \name & 27.66 & 0.873 & 0.128 \\ 
    \hline
    \end{tabular}
}
\label{table:training_quality}
\end{table}

\subsection{Training Quality Impact of \name}
\label{sec:correctness} 
The only approximation in \name is ignoring the $\epsilon$ term in the \textit{deferred optimizer update} for factoring out momentum and variance terms. To analyze its impact, we compare the rendering quality of models trained with the original method and with \name. 
Table~\ref{table:training_quality} shows this approximation has negligible impact on rendering quality, confirming that \name maintains the rendering quality of the trained models as in the original training pipeline.

\subsection{Improved Scalability and Rendering Quality}
Leveraging its memory savings discussed in Section~\ref{sec:memory_savings}, \name enables training with substantially more Gaussians under the same GPU memory budget, leading to higher rendering quality. We assess this by examining rendering quality changes with increasing Gaussian counts. Figure~\ref{fig:scalability} demonstrates that more Gaussians consistently yield higher PSNR and SSIM and lower LPIPS, indicating better rendering and reconstruction quality. The figure also shows \name extends the maximum Gaussians scaling across different platforms and systems. On a laptop with RTX 4070 Mobile GPU, \name scales the number of Gaussians from 4 million to 18 million, achieving geomean 2.6\% PSNR and 5.1\% SSIM increases, and a 28.7\% LPIPS decrease. On a desktop with RTX 4080 Super GPU, it scales the number of Gaussians from 9 million to 40 million, resulting in geomean 1.6\% PSNR and 3.6\% SSIM increases, and 30.5\% LPIPS decrease. These results substantiate that the scalability of \name directly translates into higher rendering quality across platforms.

\begin{figure}[t]
  \centering
  \includegraphics[width=\columnwidth]{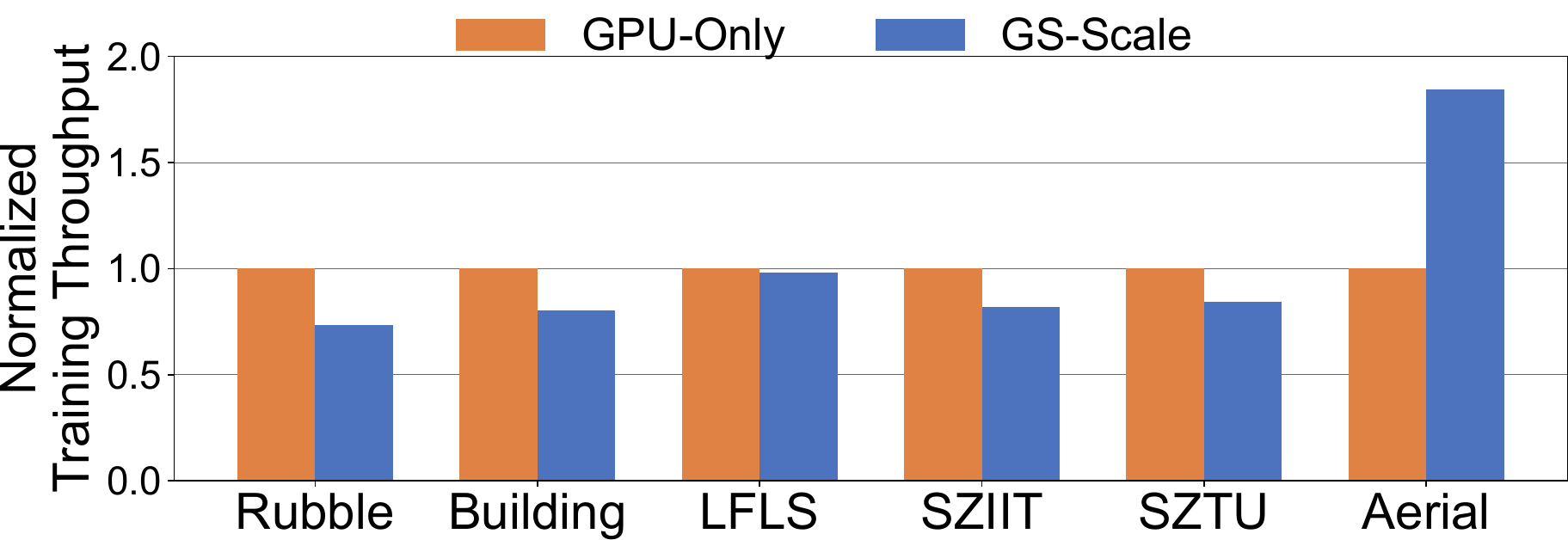}
  \caption{Training Throughput on Server Platform.}
  \label{fig:server_speedup}
\end{figure}

\subsection{Evaluation on Server Platform}
Although \name is primarily designed for laptop and desktop platforms, we also evaluate it on server platform with H100 GPU to demonstrate its broader applicability. The results on server shown in Figure~\ref{fig:server_speedup} follows a similar trend with laptop and desktop platforms, while substantial speedup is achieved on Aerial scene thanks to the large speedup gain from \textit{deferred optimizer update} as discussed in Section~\ref{sec:throughput}. We also observe that the overall training throughput normalized to GPU-only on the server is relatively lower than that of laptop despite having a similar $R_{bw}$ value. This is because the server consists of two NUMA nodes, making it relatively harder for \textit{deferred optimizer update} with random memory accesses to exploit the peak CPU memory bandwidth compared to laptop with single node. 

\subsection{Sensitivity Study}
\subsubsection*{Sensitivity to mem\_limit}
Figure~\ref{fig:mem_limit}a and b demonstrate how GPU memory usage and training throughput changes with varying \texttt{mem\_limit} thresholds. We can save more GPU memory by decreasing this threshold at the cost of slower training throughput. Slowdown results from the additional frustum culling and gradient accumulation required by image splitting. In our experiments, we use \texttt{mem\_limit} of 0.3 to prioritize training throughput over memory savings.

\subsubsection*{Sensitivity to GPU}
\begin{figure}
  \centering
  \includegraphics[width=\columnwidth]{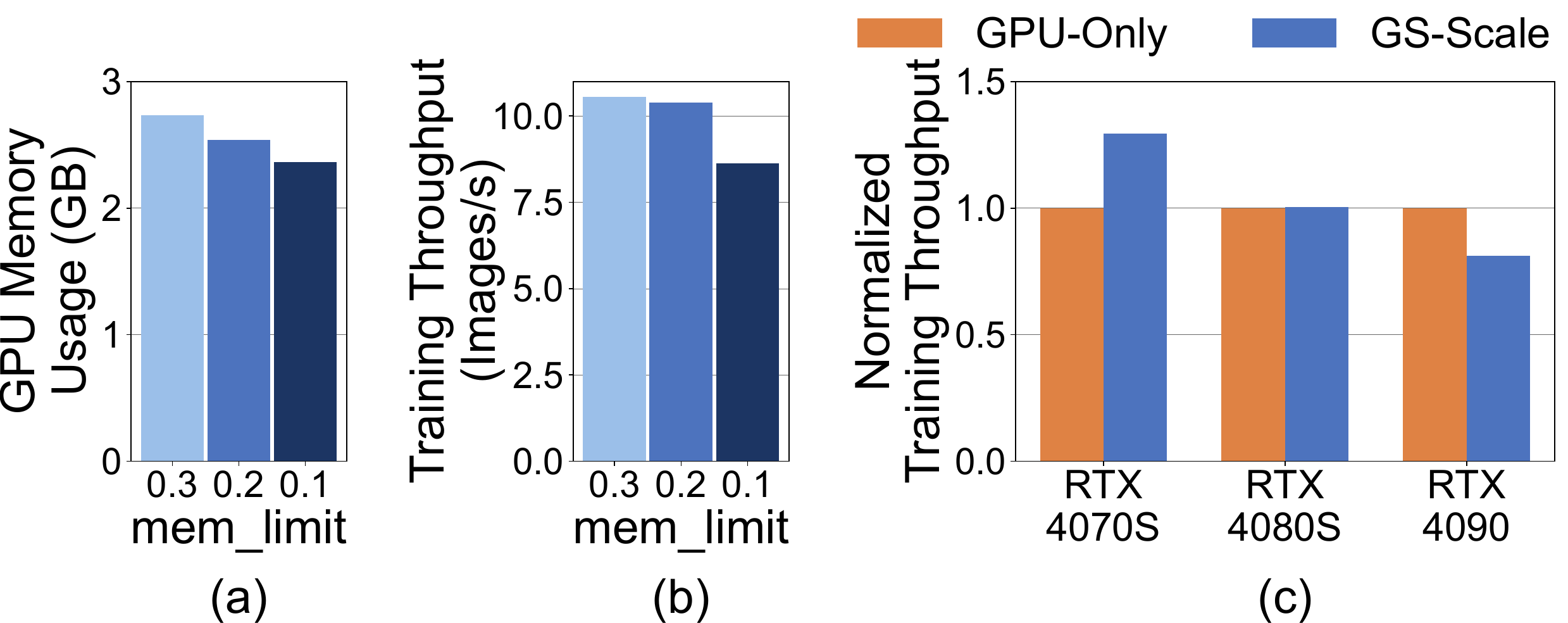}
  \caption{(a), (b) Sensitivity to \texttt{mem\_limit} on Rubble scene. (c) Sensitivity to GPU on LFLS scene. Desktop is used.}
  \label{fig:mem_limit}
\end{figure}

Figure~\ref{fig:mem_limit}c shows \name's training throughput on additional desktop GPUs (RTX 4070 Super, RTX 4090). Higher $R_{bw}$ values on RTX 4090 ($R_{bw}=11.3$ VS. $R_{bw}=5.6$ on RTX 4070 Super) with greater GPU memory bandwidth (1.01 TB/s VS. 504.2 GB/s) explains its lower normalized throughput of \name compared to GPU-Only.

\subsubsection*{Sensitivity to Image Resolution}
\begin{figure}
  \centering
  \includegraphics[width=0.95\columnwidth]{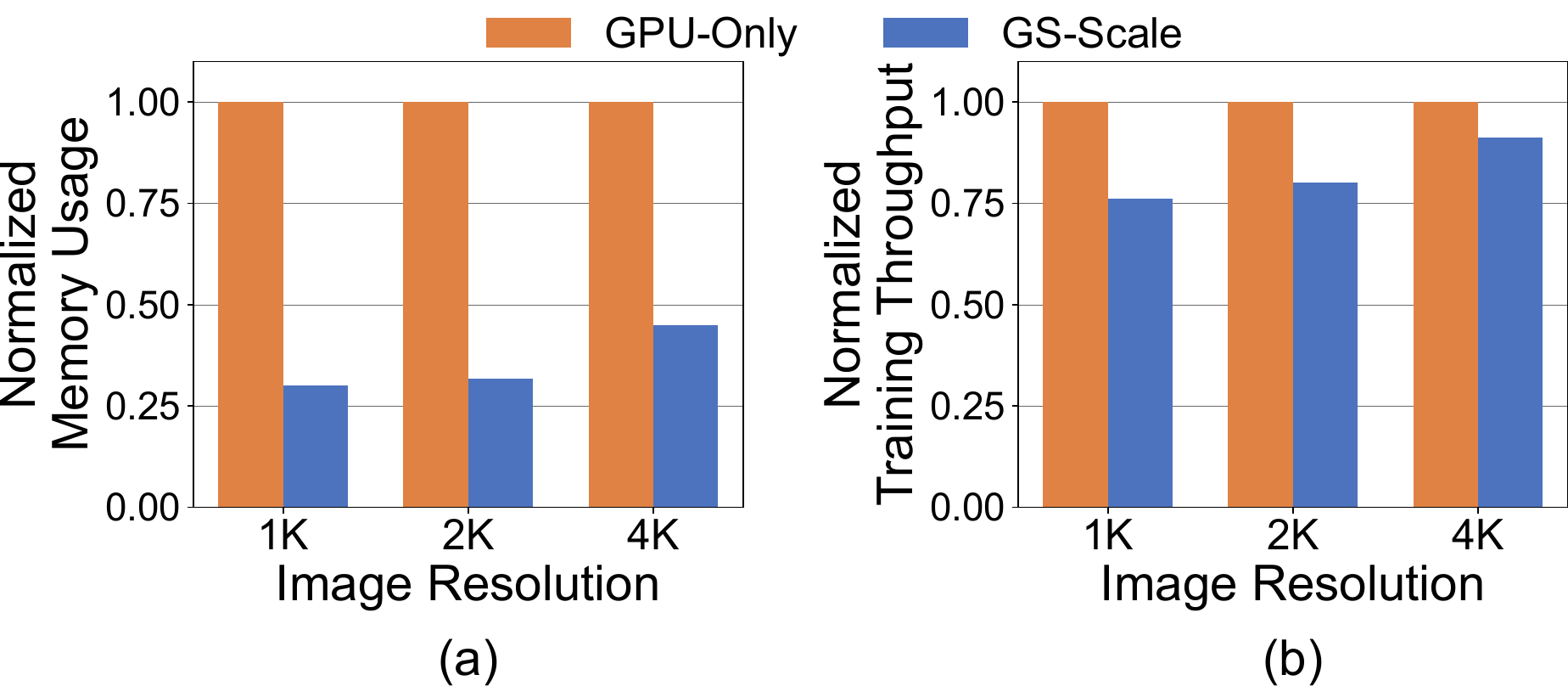}
  \caption{Impact of Image Resolution on Memory Usage and Throughput on Desktop for Rubble Scene.} 
  \label{fig:image_resolution}
\end{figure}

Figure~\ref{fig:image_resolution} shows that GPU memory savings slightly decrease as training image resolution increases since growing activation memory (Figure~\ref{fig:breakdown_memory}) reduces the relative portion of offloadable parameters, optimizer states, and gradients.
Conversely, relative training throughput increases. This is because higher resolutions slow down the GPU-based forward/backward pass, providing more slack for pipelining CPU-based optimizer updates.

\section{Related Work}
\label{sec:related}
\subsubsection*{Acceleration on 3D Gaussian Splatting Rendering.}  
Several works have been proposed to accelerate 3DGS rendering through both software optimizations~\cite{flashgs, gscache, octreegs, progs, eagles, lightgaussian, pup3dgs, sortfreegs} and specialized accelerators~\cite{gscore, lumina, metasapiens, lsgaussian, gbu, uni}. For software-only solutions,
GS-Cache~\cite{gscache} reduces redundant computations via caching data across frames, combined with an efficient scheduler and optimized GPU kernels. 
FlashGS~\cite{flashgs} eliminates unnecessary computations via a precise intersection test and improves GPU utilization by overlapping memory access with computation.
For hardware accelerators, GSCore~\cite{gscore} introduces the first dedicated accelerator for 3DGS, eliminating sorting and rasterization for unnecessary Gaussians through algorithm-hardware co-design.
Lumina~\cite{lumina} also mitigates sorting and rasterization bottlenecks by sharing sorting results across frames and caching previous rendering results via hardware support.
MetaSapiens~\cite{metasapiens} adopts efficiency-aware pruning and foveated rendering, co-designed with a specialized accelerator to enable real-time rendering.

\subsubsection*{Acceleration on 3D Gaussian Splatting Training.}
3DGS training suffers from large amount of atomic operations during gradient accumulation and various works~\cite{distwar, arc, gsarch, taming3dgs, litegs} have been proposed to address this bottleneck.
DISTWAR~\cite{distwar} accelerates atomic operations by enabling warp-level reduction and leveraging L2 atomic units (ROP units). 
ARC~\cite{arc} further addresses atomic bottlenecks by introducing specialized hardware unit for atomic operations.
GSArch~\cite{gsarch} reduces both off-chip and on-chip memory accesses in 3DGS training through gradient pruning and on-chip memory access rearrangement.

\subsubsection*{Scaling 3D Gaussian Splatting}
Recent works have tackeled the challenges in large scale 3DGS training via both algorithm level and system level solutions.
Most algorithm-level approaches~\cite{vastgaussian, retinags, citygaussian, citygaussianv2, hierarchicalgs, hug} follow a divide-and-conquer strategy: partitioning the 3D scene into smaller chunks, training them independently, and later merging the results. While this avoids out-of-memory errors, it fundamentally alters the original 3DGS training recipe.
Grendel~\cite{grendel} is the first framework to enable large-scale 3DGS training without modifying the original 3DGS algorithm. By addressing GPU load imbalance and inter-GPU communication overhead in a distributed setting, Grendel supports efficient training with tens of millions of Gaussians. Importantly, Grendel shows that simply supporting the original 3DGS pipeline alone leads to substantially faster training and superior rendering quality compared to divide-and-conquer methods, highlighting the importance of system-level solutions in 3DGS training.
\section{Conclusion}
\label{sec:con}
3D Gaussian Splatting offers high visual quality and fast rendering speed, but its training demands significant GPU memory. \name resolves this by offloading Gaussians to host memory, transferring only necessary subsets to the GPU on demand, greatly reducing GPU memory usage. \name also optimizes CPU-based frustum culling and optimizer updates through selective offloading, parameter forwarding, and a deferred optimizer update. Experiments show \name saves GPU memory demands by 3.3$\times$-5.6$\times$, maintaining training throughput comparable to GPU-only systems. This enables \name to facilitate much larger-scale 3DGS training on commodity GPUs, achieving geomean 28.7\% and 30.5\% LPIPS improvement on an RTX 4070 Mobile GPU and RTX 4080 Super GPU respectively.


\bibliographystyle{ACM-Reference-Format}
\bibliography{main.bib}


\begin{thebibliography}{54}


\ifx \showCODEN    \undefined \def \showCODEN     #1{\unskip}     \fi
\ifx \showISBNx    \undefined \def \showISBNx     #1{\unskip}     \fi
\ifx \showISBNxiii \undefined \def \showISBNxiii  #1{\unskip}     \fi
\ifx \showISSN     \undefined \def \showISSN      #1{\unskip}     \fi
\ifx \showLCCN     \undefined \def \showLCCN      #1{\unskip}     \fi
\ifx \shownote     \undefined \def \shownote      #1{#1}          \fi
\ifx \showarticletitle \undefined \def \showarticletitle #1{#1}   \fi
\ifx \showURL      \undefined \def \showURL       {\relax}        \fi
\providecommand\bibfield[2]{#2}
\providecommand\bibinfo[2]{#2}
\providecommand\natexlab[1]{#1}
\providecommand\showeprint[2][]{arXiv:#2}

\bibitem[ark(2022)]%
        {arkio}
 \bibinfo{year}{2022}\natexlab{}.
\newblock \bibinfo{title}{Design better buildings, together}.
\newblock \bibinfo{howpublished}{\url{https://www.arkio.is}}.
\newblock


\bibitem[ske(2022)]%
        {sketchfab}
 \bibinfo{year}{2022}\natexlab{}.
\newblock \bibinfo{title}{Metaverse 3D models}.
\newblock \bibinfo{howpublished}{\url{https://sketchfab.com/tags/metaverse}}.
\newblock


\bibitem[pol(2023)]%
        {polycam}
 \bibinfo{year}{2023}\natexlab{}.
\newblock \bibinfo{title}{3D Gaussian Splatting: Create and view splats for free}.
\newblock \bibinfo{howpublished}{\url{https://poly.cam/tools/gaussian-splatting}}.
\newblock


\bibitem[lap(2023)]%
        {laptop}
 \bibinfo{year}{2023}\natexlab{}.
\newblock \bibinfo{title}{ASUS TUF Gaming F17}.
\newblock \bibinfo{howpublished}{\url{https://www.asus.com/laptops/for-gaming/tuf-gaming/asus-tuf-gaming-f17-2023}}.
\newblock


\bibitem[rea(2025)]%
        {realhorizon}
 \bibinfo{year}{2025}\natexlab{}.
\newblock \bibinfo{title}{Virtual Tours: Explore properties and spaces from anywhere in the world}.
\newblock \bibinfo{howpublished}{\url{https://www.realhorizons.in/tours}}.
\newblock


\bibitem[sca(2025)]%
        {scaniverse}
 \bibinfo{year}{2025}\natexlab{}.
\newblock \bibinfo{title}{Your invitation to explore the world in 3D}.
\newblock \bibinfo{howpublished}{\url{https://scaniverse.com}}.
\newblock


\bibitem[Chen et~al\mbox{.}(2022)]%
        {tensorf}
\bibfield{author}{\bibinfo{person}{Anpei Chen}, \bibinfo{person}{Zexiang Xu}, \bibinfo{person}{Andreas Geiger}, \bibinfo{person}{Jingyi Yu}, {and} \bibinfo{person}{Hao Su}.} \bibinfo{year}{2022}\natexlab{}.
\newblock \showarticletitle{TensoRF: Tensorial Radiance Fields}. In \bibinfo{booktitle}{\emph{European Conference on Computer Vision (ECCV)}}.
\newblock


\bibitem[Chen et~al\mbox{.}(2025)]%
        {taoavatar}
\bibfield{author}{\bibinfo{person}{Jianchuan Chen}, \bibinfo{person}{Jingchuan Hu}, \bibinfo{person}{Gaige Wang}, \bibinfo{person}{Zhonghua Jiang}, \bibinfo{person}{Tiansong Zhou}, \bibinfo{person}{Zhiwen Chen}, {and} \bibinfo{person}{Chengfei Lv}.} \bibinfo{year}{2025}\natexlab{}.
\newblock \showarticletitle{TaoAvatar: Real-Time Lifelike Full-Body Talking Avatars for Augmented Reality via 3D Gaussian Splatting}. In \bibinfo{booktitle}{\emph{IEEE/CVF Conference on Computer Vision and Pattern Recognition (CVPR)}}.
\newblock


\bibitem[Deng et~al\mbox{.}(2022)]%
        {fovnerf}
\bibfield{author}{\bibinfo{person}{Nianchen Deng}, \bibinfo{person}{Zhenyi He}, \bibinfo{person}{Jiannan Ye}, \bibinfo{person}{Budmonde Duinkharjav}, \bibinfo{person}{Praneeth Chakravarthula}, \bibinfo{person}{Xubo Yang}, {and} \bibinfo{person}{Qi Sun}.} \bibinfo{year}{2022}\natexlab{}.
\newblock \showarticletitle{FoV-NeRF: Foveated Neural Radiance Fields for Virtual Reality}. In \bibinfo{booktitle}{\emph{IEEE Transactions on Visualization and Computer Graphics}}.
\newblock


\bibitem[Durvasula et~al\mbox{.}(2025)]%
        {arc}
\bibfield{author}{\bibinfo{person}{Sankeerth Durvasula}, \bibinfo{person}{Adrian Zhao}, \bibinfo{person}{Fan Chen}, \bibinfo{person}{Ruofan Liang}, \bibinfo{person}{Pawan~Kumar Sanjaya}, \bibinfo{person}{Yushi Guan}, \bibinfo{person}{Christina Giannoula}, {and} \bibinfo{person}{Nandita Vijaykumar}.} \bibinfo{year}{2025}\natexlab{}.
\newblock \showarticletitle{ARC: Warp-level Adaptive Atomic Reduction in GPUs to Accelerate Differentiable Rendering}. In \bibinfo{booktitle}{\emph{ACM International Conference on Architectural Support for Programming Languages and Operating Systems (ASPLOS)}}.
\newblock


\bibitem[Durvasula et~al\mbox{.}(2024)]%
        {distwar}
\bibfield{author}{\bibinfo{person}{Sankeerth Durvasula}, \bibinfo{person}{Adrian Zhao}, \bibinfo{person}{Fan Chen}, \bibinfo{person}{Ruofan Liang}, \bibinfo{person}{Pawan~Kumar Sanjaya}, {and} \bibinfo{person}{Nandita Vijaykumar}.} \bibinfo{year}{2024}\natexlab{}.
\newblock \showarticletitle{DISTWAR: Fast Differentiable Rendering on Raster-based Rendering Pipelines}.
\newblock \bibinfo{journal}{\emph{arXiv preprint arXiv:2401.05345}} (\bibinfo{year}{2024}).
\newblock


\bibitem[Fan et~al\mbox{.}(2024)]%
        {lightgaussian}
\bibfield{author}{\bibinfo{person}{Zhiwen Fan}, \bibinfo{person}{Kevin Wang}, \bibinfo{person}{Kairun Wen}, \bibinfo{person}{Zehao Zhu}, \bibinfo{person}{Dejia Xu}, {and} \bibinfo{person}{Zhangyang Wang}.} \bibinfo{year}{2024}\natexlab{}.
\newblock \showarticletitle{LightGaussian: Unbounded 3D Gaussian Compression with 15× Reduction and 200+ FPS}. In \bibinfo{booktitle}{\emph{Advances in Neural Information Processing Systems (NeurIPS)}}.
\newblock


\bibitem[Feng et~al\mbox{.}(2025a)]%
        {flashgs}
\bibfield{author}{\bibinfo{person}{Guofeng Feng}, \bibinfo{person}{Siyan Chen}, \bibinfo{person}{Rong Fu}, \bibinfo{person}{Zimu Liao}, \bibinfo{person}{Yi Wang}, \bibinfo{person}{Tao Liu}, \bibinfo{person}{Boni Hu}, \bibinfo{person}{Linning Xu}, \bibinfo{person}{Zhilin Pei}, \bibinfo{person}{Hengjie Li}, \bibinfo{person}{Xiuhong Li}, \bibinfo{person}{Ninghui Sun}, \bibinfo{person}{Xingcheng Zhang}, {and} \bibinfo{person}{Bo Dai}.} \bibinfo{year}{2025}\natexlab{a}.
\newblock \showarticletitle{FlashGS: Efficient 3D Gaussian Splatting for Large-scale and High-resolution Rendering}. In \bibinfo{booktitle}{\emph{IEEE/CVF Conference on Computer Vision and Pattern Recognition (CVPR)}}.
\newblock


\bibitem[Feng et~al\mbox{.}(2025b)]%
        {lumina}
\bibfield{author}{\bibinfo{person}{Yu Feng}, \bibinfo{person}{Weikai Lin}, \bibinfo{person}{Yuge Cheng}, \bibinfo{person}{Zihan Liu}, \bibinfo{person}{Jingwen Leng}, \bibinfo{person}{Minyi Guo}, \bibinfo{person}{Chen Chen}, \bibinfo{person}{Shixuan Sun}, {and} \bibinfo{person}{Yuhao Zhu}.} \bibinfo{year}{2025}\natexlab{b}.
\newblock \showarticletitle{Lumina: Real-Time Neural Rendering by Exploiting Computational Redundancy}. In \bibinfo{booktitle}{\emph{ACM/IEEE International Symposium on Computer Architecture (ISCA)}}.
\newblock


\bibitem[Franke et~al\mbox{.}(2025)]%
        {vrsplatting}
\bibfield{author}{\bibinfo{person}{Linus Franke}, \bibinfo{person}{Laura Fink}, {and} \bibinfo{person}{Marc Stamminger}.} \bibinfo{year}{2025}\natexlab{}.
\newblock \showarticletitle{VR-Splatting: Foveated Radiance Field Rendering via 3D Gaussian Splatting and Neural Points}.
\newblock \bibinfo{journal}{\emph{Proc. ACM Comput. Graph. Interact. Tech. (PACMCGIT)}} (\bibinfo{year}{2025}).
\newblock


\bibitem[Garbin et~al\mbox{.}(2021)]%
        {fastnerf}
\bibfield{author}{\bibinfo{person}{Stephan~J. Garbin}, \bibinfo{person}{Marek Kowalski}, \bibinfo{person}{Matthew Johnson}, \bibinfo{person}{Jamie Shotton}, {and} \bibinfo{person}{Julien Valentin}.} \bibinfo{year}{2021}\natexlab{}.
\newblock \showarticletitle{FastNeRF: High-Fidelity Neural Rendering at 200FPS}. In \bibinfo{booktitle}{\emph{IEEE/CVF International Conference on Computer Vision (ICCV)}}.
\newblock


\bibitem[Guo et~al\mbox{.}(2025)]%
        {articulatedgs}
\bibfield{author}{\bibinfo{person}{Junfu Guo}, \bibinfo{person}{Yu Xin}, \bibinfo{person}{Gaoyi Liu}, \bibinfo{person}{Kai Xu}, \bibinfo{person}{Ligang Liu}, {and} \bibinfo{person}{Ruizhen Hu}.} \bibinfo{year}{2025}\natexlab{}.
\newblock \showarticletitle{ArticulatedGS: Self-supervised Digital Twin Modeling of Articulated Objects using 3D Gaussian Splatting}. In \bibinfo{booktitle}{\emph{IEEE/CVF Conference on Computer Vision and Pattern Recognition (CVPR)}}.
\newblock


\bibitem[Hanson et~al\mbox{.}(2025)]%
        {pup3dgs}
\bibfield{author}{\bibinfo{person}{Alex Hanson}, \bibinfo{person}{Allen Tu}, \bibinfo{person}{Vasu Singla}, \bibinfo{person}{Mayuka Jayawardhana}, \bibinfo{person}{Matthias Zwicker}, {and} \bibinfo{person}{Tom Goldstein}.} \bibinfo{year}{2025}\natexlab{}.
\newblock \showarticletitle{PUP 3D-GS: Principled Uncertainty Pruning for 3D Gaussian Splatting}. In \bibinfo{booktitle}{\emph{IEEE/CVF Conference on Computer Vision and Pattern Recognition (CVPR)}}.
\newblock


\bibitem[He et~al\mbox{.}(2025)]%
        {gsarch}
\bibfield{author}{\bibinfo{person}{Houshu He}, \bibinfo{person}{Gang Li}, \bibinfo{person}{Fangxin Liu}, \bibinfo{person}{Li Jiang}, \bibinfo{person}{Xiaoyao Liang}, {and} \bibinfo{person}{Zhuoran Song}.} \bibinfo{year}{2025}\natexlab{}.
\newblock \showarticletitle{GSArch: Breaking Memory Barriers in 3D Gaussian Splatting Training via Architectural Support}. In \bibinfo{booktitle}{\emph{IEEE International Symposium on High Performance Computer Architecture (HPCA)}}.
\newblock


\bibitem[Hou et~al\mbox{.}(2025)]%
        {sortfreegs}
\bibfield{author}{\bibinfo{person}{Qiqi Hou}, \bibinfo{person}{Randall Rauwendaal}, \bibinfo{person}{Zifeng Li}, \bibinfo{person}{Hoang Le}, \bibinfo{person}{Farzad Farhadzadeh}, \bibinfo{person}{Fatih Porikli}, \bibinfo{person}{Alexei Bourd}, {and} \bibinfo{person}{Amir Said}.} \bibinfo{year}{2025}\natexlab{}.
\newblock \showarticletitle{Sort-free Gaussian Splatting via Weighted Sum Rendering}. In \bibinfo{booktitle}{\emph{International Conference on Learning Representations (ICLR)}}.
\newblock


\bibitem[Jiang et~al\mbox{.}(2024)]%
        {vrgs}
\bibfield{author}{\bibinfo{person}{Ying Jiang}, \bibinfo{person}{Chang Yu}, \bibinfo{person}{Tianyi Xie}, \bibinfo{person}{Xuan Li}, \bibinfo{person}{Yutao Feng}, \bibinfo{person}{Huamin Wang}, \bibinfo{person}{Minchen Li}, \bibinfo{person}{Henry Lau}, \bibinfo{person}{Feng Gao}, \bibinfo{person}{Yin Yang}, {and} \bibinfo{person}{Chenfanfu Jiang}.} \bibinfo{year}{2024}\natexlab{}.
\newblock \showarticletitle{VR-GS: A Physical Dynamics-Aware Interactive Gaussian Splatting System in Virtual Reality}. In \bibinfo{booktitle}{\emph{ACM Transactions on Graphics (SIGGRAPH)}}.
\newblock


\bibitem[Kerbl et~al\mbox{.}(2023)]%
        {3dgs}
\bibfield{author}{\bibinfo{person}{Bernhard Kerbl}, \bibinfo{person}{Georgios Kopanas}, \bibinfo{person}{Thomas Leimkühler}, {and} \bibinfo{person}{George Drettakis}.} \bibinfo{year}{2023}\natexlab{}.
\newblock \showarticletitle{3D Gaussian Splatting for Real-Time Radiance Field Rendering}. In \bibinfo{booktitle}{\emph{ACM Transactions on Graphics (SIGGRAPH)}}.
\newblock


\bibitem[Kerbl et~al\mbox{.}(2024)]%
        {hierarchicalgs}
\bibfield{author}{\bibinfo{person}{Bernhard Kerbl}, \bibinfo{person}{Andréas Meuleman}, \bibinfo{person}{Georgios Kopanas}, \bibinfo{person}{Michael Wimmer}, \bibinfo{person}{Alexandre Lanvin}, {and} \bibinfo{person}{George Drettakis}.} \bibinfo{year}{2024}\natexlab{}.
\newblock \showarticletitle{A Hierarchical 3D Gaussian Representation for Real-Time Rendering of Very Large Datasets}. In \bibinfo{booktitle}{\emph{ACM Transactions on Graphics (SIGGRAPH)}}.
\newblock


\bibitem[Kingma and Ba(2015)]%
        {adam}
\bibfield{author}{\bibinfo{person}{Diederik~P. Kingma} {and} \bibinfo{person}{Jimmy Ba}.} \bibinfo{year}{2015}\natexlab{}.
\newblock \showarticletitle{Adam: A Method for Stochastic Optimization}. In \bibinfo{booktitle}{\emph{International Conference on Learning Representations (ICLR)}}.
\newblock


\bibitem[Lee et~al\mbox{.}(2024)]%
        {gscore}
\bibfield{author}{\bibinfo{person}{Junseo Lee}, \bibinfo{person}{Seokwon Lee}, \bibinfo{person}{Jungi Lee}, \bibinfo{person}{Junyong Park}, {and} \bibinfo{person}{Jaewoong Sim}.} \bibinfo{year}{2024}\natexlab{}.
\newblock \showarticletitle{GSCore: Efficient Radiance Field Rendering via Architectural Support for 3D Gaussian Splatting}. In \bibinfo{booktitle}{\emph{ACM International Conference on Architectural Support for Programming Languages and Operating Systems}}.
\newblock


\bibitem[Li et~al\mbox{.}(2024)]%
        {retinags}
\bibfield{author}{\bibinfo{person}{Bingling Li}, \bibinfo{person}{Shengyi Chen}, \bibinfo{person}{Luchao Wang}, \bibinfo{person}{Kaimin Liao}, \bibinfo{person}{Sijie Yan}, {and} \bibinfo{person}{Yuanjun Xiong}.} \bibinfo{year}{2024}\natexlab{}.
\newblock \showarticletitle{RetinaGS: Scalable Training for Dense Scene Rendering with Billion-Scale 3D Gaussians}. In \bibinfo{booktitle}{\emph{arXiv preprint arXiv:2406.11836}}.
\newblock


\bibitem[Li et~al\mbox{.}(2025)]%
        {uni}
\bibfield{author}{\bibinfo{person}{Chaojian Li}, \bibinfo{person}{Sixu Li}, \bibinfo{person}{Linrui Jiang}, \bibinfo{person}{Jingqun Zhang}, {and} \bibinfo{person}{Yingyan~Celine Lin}.} \bibinfo{year}{2025}\natexlab{}.
\newblock \showarticletitle{Uni-Render: A Unified Accelerator for Real-Time Rendering Across Diverse Neural Renderers}. In \bibinfo{booktitle}{\emph{IEEE International Symposium on High Performance Computer Architecture (HPCA)}}.
\newblock


\bibitem[Li et~al\mbox{.}(2023)]%
        {matrixcity}
\bibfield{author}{\bibinfo{person}{Yixuan Li}, \bibinfo{person}{Lihan Jiang}, \bibinfo{person}{Linning Xu}, \bibinfo{person}{Yuanbo Xiangli}, \bibinfo{person}{Zhenzhi Wang}, \bibinfo{person}{Dahua Lin}, {and} \bibinfo{person}{Bo Dai}.} \bibinfo{year}{2023}\natexlab{}.
\newblock \showarticletitle{MatrixCity: A Large-scale City Dataset for City-scale Neural Rendering and Beyond}. In \bibinfo{booktitle}{\emph{IEEE/CVF International Conference on Computer Vision (ICCV)}}.
\newblock


\bibitem[Liao et~al\mbox{.}(2025)]%
        {litegs}
\bibfield{author}{\bibinfo{person}{Kaimin Liao}, \bibinfo{person}{Hua Wang}, \bibinfo{person}{Zhi Chen}, \bibinfo{person}{Luchao Wang}, {and} \bibinfo{person}{Yaohua Tang}.} \bibinfo{year}{2025}\natexlab{}.
\newblock \showarticletitle{LiteGS: A High-performance Framework to Train 3DGS in Subminutes via System and Algorithm Codesign}.
\newblock \bibinfo{journal}{\emph{arXiv preprint arXiv:2503.01199}} (\bibinfo{year}{2025}).
\newblock


\bibitem[Liao et~al\mbox{.}(2024)]%
        {eagles}
\bibfield{author}{\bibinfo{person}{Zhimeng Liao}, \bibinfo{person}{Xinyang Li}, \bibinfo{person}{Shaohui Liu}, \bibinfo{person}{Jiakai Zhang}, \bibinfo{person}{Xian Liu}, \bibinfo{person}{Yikai Wang}, \bibinfo{person}{Ying Feng}, \bibinfo{person}{Xiaoxiao Long}, \bibinfo{person}{Shuguang Cui}, {and} \bibinfo{person}{Wenping Wang}.} \bibinfo{year}{2024}\natexlab{}.
\newblock \showarticletitle{EAGLES: Efficient Accelerated 3D Gaussians with Lightweight Encodings}. In \bibinfo{booktitle}{\emph{European Conference on Computer Vision (ECCV)}}.
\newblock


\bibitem[Lin et~al\mbox{.}(2024)]%
        {vastgaussian}
\bibfield{author}{\bibinfo{person}{Jiaqi Lin}, \bibinfo{person}{Zhihao Li}, \bibinfo{person}{Xiao Tang}, \bibinfo{person}{Jianzhuang Liu}, \bibinfo{person}{Shiyong Liu}, \bibinfo{person}{Jiayue Liu}, \bibinfo{person}{Yangdi Lu}, \bibinfo{person}{Xiaofei Wu}, \bibinfo{person}{Songcen Xu}, \bibinfo{person}{Youliang Yan}, {and} \bibinfo{person}{Wenming Yang}.} \bibinfo{year}{2024}\natexlab{}.
\newblock \showarticletitle{VastGaussian: Vast 3D Gaussians for Large Scene Reconstruction}. In \bibinfo{booktitle}{\emph{IEEE/CVF Conference on Computer Vision and Pattern Recognition (CVPR)}}.
\newblock


\bibitem[Lin et~al\mbox{.}(2025)]%
        {metasapiens}
\bibfield{author}{\bibinfo{person}{Weikai Lin}, \bibinfo{person}{Yu Feng}, {and} \bibinfo{person}{Yuhao Zhu}.} \bibinfo{year}{2025}\natexlab{}.
\newblock \showarticletitle{MetaSapiens: Real-Time Neural Rendering with Efficiency-Aware Pruning and Accelerated Foveated Rendering}. In \bibinfo{booktitle}{\emph{International Conference on Architectural Support for Programming Languages and Operating Systems (ASPLOS)}}.
\newblock


\bibitem[Liu et~al\mbox{.}(2024)]%
        {citygaussian}
\bibfield{author}{\bibinfo{person}{Yang Liu}, \bibinfo{person}{He Guan}, \bibinfo{person}{Chuanchen Luo}, \bibinfo{person}{Lue Fan}, \bibinfo{person}{Naiyan Wang}, \bibinfo{person}{Junran Peng}, {and} \bibinfo{person}{Zhaoxiang Zhang}.} \bibinfo{year}{2024}\natexlab{}.
\newblock \showarticletitle{CityGaussian: Real-time High-quality Large-Scale Scene Rendering with Gaussians}. In \bibinfo{booktitle}{\emph{European Conference on Computer Vision (ECCV)}}.
\newblock


\bibitem[Liu et~al\mbox{.}(2025)]%
        {citygaussianv2}
\bibfield{author}{\bibinfo{person}{Yang Liu}, \bibinfo{person}{Chuanchen Luo}, \bibinfo{person}{Zhongkai Mao}, \bibinfo{person}{Junran Peng}, {and} \bibinfo{person}{Zhaoxiang Zhang}.} \bibinfo{year}{2025}\natexlab{}.
\newblock \showarticletitle{CityGaussianV2: Efficient and Geometrically Accurate Reconstruction for Large-Scale Scenes}. In \bibinfo{booktitle}{\emph{International Conference on Learning Representations (ICLR)}}.
\newblock


\bibitem[Loshchilov and Hutter(2019)]%
        {adamw}
\bibfield{author}{\bibinfo{person}{Ilya Loshchilov} {and} \bibinfo{person}{Frank Hutter}.} \bibinfo{year}{2019}\natexlab{}.
\newblock \showarticletitle{Decoupled Weight Decay Regularization}. In \bibinfo{booktitle}{\emph{International Conference on Learning Representations (ICLR)}}.
\newblock


\bibitem[Mallick et~al\mbox{.}(2024)]%
        {taming3dgs}
\bibfield{author}{\bibinfo{person}{Saswat~Subhajyoti Mallick}, \bibinfo{person}{Rahul Goel}, \bibinfo{person}{Bernhard Kerbl}, \bibinfo{person}{Francisco~Vicente Carrasco}, \bibinfo{person}{Markus Steinberger}, {and} \bibinfo{person}{Fernando De~La Torre}.} \bibinfo{year}{2024}\natexlab{}.
\newblock \showarticletitle{Taming 3DGS: High-Quality Radiance Fields with Limited Resources}. In \bibinfo{booktitle}{\emph{SIGGRAPH Asia 2024 Conference Papers}}.
\newblock


\bibitem[Mildenhall et~al\mbox{.}(2020)]%
        {nerf}
\bibfield{author}{\bibinfo{person}{Ben Mildenhall}, \bibinfo{person}{Pratul~P. Srinivasan}, \bibinfo{person}{Matthew Tancik}, \bibinfo{person}{Jonathan~T. Barron}, \bibinfo{person}{Ravi Ramamoorthi}, {and} \bibinfo{person}{Ren Ng}.} \bibinfo{year}{2020}\natexlab{}.
\newblock \showarticletitle{NeRF: Representing Scenes as Neural Radiance Fields for View Synthesis}. In \bibinfo{booktitle}{\emph{European Conference on Computer Vision (ECCV)}}.
\newblock


\bibitem[M\"uller et~al\mbox{.}(2022)]%
        {instantngp}
\bibfield{author}{\bibinfo{person}{Thomas M\"uller}, \bibinfo{person}{Alex Evans}, \bibinfo{person}{Christoph Schied}, {and} \bibinfo{person}{Alexander Keller}.} \bibinfo{year}{2022}\natexlab{}.
\newblock \showarticletitle{Instant Neural Graphics Primitives with a Multiresolution Hash Encoding}. In \bibinfo{booktitle}{\emph{ACM Transactions on Graphics (SIGGRAPH)}}.
\newblock


\bibitem[Park et~al\mbox{.}(2025)]%
        {decdec}
\bibfield{author}{\bibinfo{person}{Yeonhong Park}, \bibinfo{person}{Jake Hyun}, \bibinfo{person}{Hojoon Kim}, {and} \bibinfo{person}{Jae~W. Lee}.} \bibinfo{year}{2025}\natexlab{}.
\newblock \showarticletitle{DecDEC: A Systems Approach to Advancing Low-Bit LLM Quantization}. In \bibinfo{booktitle}{\emph{USENIX Symposium on Operating Systems Design and Implementation (OSDI)}}.
\newblock


\bibitem[Paszke et~al\mbox{.}(2019)]%
        {pytorch}
\bibfield{author}{\bibinfo{person}{Adam Paszke}, \bibinfo{person}{Sam Gross}, \bibinfo{person}{Francisco Massa}, \bibinfo{person}{Adam Lerer}, \bibinfo{person}{James Bradbury}, \bibinfo{person}{Gregory Chanan}, \bibinfo{person}{Trevor Killeen}, \bibinfo{person}{Zeming Lin}, \bibinfo{person}{Natalia Gimelshein}, \bibinfo{person}{Luca Antiga}, \bibinfo{person}{Alban Desmaison}, \bibinfo{person}{Andreas Kopf}, \bibinfo{person}{Edward Yang}, \bibinfo{person}{Zachary DeVito}, \bibinfo{person}{Martin Raison}, \bibinfo{person}{Alykhan Tejani}, \bibinfo{person}{Sasank Chilamkurthy}, \bibinfo{person}{Benoit Steiner}, \bibinfo{person}{Lu Fang}, \bibinfo{person}{Junjie Bai}, {and} \bibinfo{person}{Soumith Chintala}.} \bibinfo{year}{2019}\natexlab{}.
\newblock \showarticletitle{PyTorch: An Imperative Style, High-Performance Deep Learning Library}. In \bibinfo{booktitle}{\emph{Advances in Neural Information Processing Systems (NeurIPS)}}.
\newblock


\bibitem[Ren et~al\mbox{.}(2025)]%
        {octreegs}
\bibfield{author}{\bibinfo{person}{Kerui Ren}, \bibinfo{person}{Lihan Jiang}, \bibinfo{person}{Tao Lu}, \bibinfo{person}{Mulin Yu}, \bibinfo{person}{Linning Xu}, \bibinfo{person}{Zhangkai Ni}, {and} \bibinfo{person}{Bo Dai}.} \bibinfo{year}{2025}\natexlab{}.
\newblock \showarticletitle{Octree-GS: Towards Consistent Real-time Rendering with LOD-Structured 3D Gaussians}.
\newblock \bibinfo{journal}{\emph{IEEE Transactions on Pattern Analysis and Machine Intelligence (TPAMI)}} (\bibinfo{year}{2025}).
\newblock


\bibitem[Schneider et~al\mbox{.}(2025)]%
        {worldexplorer}
\bibfield{author}{\bibinfo{person}{Manuel-Andreas Schneider}, \bibinfo{person}{Lukas Höllein}, {and} \bibinfo{person}{Matthias Nießner}.} \bibinfo{year}{2025}\natexlab{}.
\newblock \showarticletitle{WorldExplorer: Towards Generating Fully Navigable 3D Scenes}.
\newblock \bibinfo{journal}{\emph{arXiv preprint arXiv:2506.01799}} (\bibinfo{year}{2025}).
\newblock


\bibitem[Snavely et~al\mbox{.}(2006)]%
        {sfm}
\bibfield{author}{\bibinfo{person}{Noah Snavely}, \bibinfo{person}{Steven~M. Seitz}, {and} \bibinfo{person}{Richard Szeliski}.} \bibinfo{year}{2006}\natexlab{}.
\newblock \showarticletitle{Photo tourism: exploring photo collections in 3D}. In \bibinfo{booktitle}{\emph{ACM Transactions on Graphics (SIGGRAPH)}}.
\newblock


\bibitem[Su et~al\mbox{.}(2025)]%
        {hug}
\bibfield{author}{\bibinfo{person}{Mai Su}, \bibinfo{person}{Zhongtao Wang}, \bibinfo{person}{Huishan Au}, \bibinfo{person}{Yilong Li}, \bibinfo{person}{Xizhe Cao}, \bibinfo{person}{Chengwei Pan}, \bibinfo{person}{Yisong Chen}, {and} \bibinfo{person}{Guoping Wang}.} \bibinfo{year}{2025}\natexlab{}.
\newblock \showarticletitle{HUG: Hierarchical Urban Gaussian Splatting with Block-Based Reconstruction for Large-Scale Aerial Scenes}. In \bibinfo{booktitle}{\emph{IEEE/CVF International Conference on Computer Vision (ICCV)}}.
\newblock


\bibitem[Tao et~al\mbox{.}(2025)]%
        {gscache}
\bibfield{author}{\bibinfo{person}{Miao Tao}, \bibinfo{person}{Yuanzhen Zhou}, \bibinfo{person}{Haoran Xu}, \bibinfo{person}{Zeyu He}, \bibinfo{person}{Zhenyu Yang}, \bibinfo{person}{Yuchang Zhang}, \bibinfo{person}{Zhongling Su}, \bibinfo{person}{Linning Xu}, \bibinfo{person}{Zhenxiang Ma}, \bibinfo{person}{Rong Fu}, \bibinfo{person}{Hengjie Li}, \bibinfo{person}{Xingcheng Zhang}, {and} \bibinfo{person}{Jidong Zhai}.} \bibinfo{year}{2025}\natexlab{}.
\newblock \showarticletitle{GS-Cache: A GS-Cache Inference Framework for Large-Scale Gaussian Splatting Models}.
\newblock \bibinfo{journal}{\emph{arXiv preprint arXiv:2502.14938}} (\bibinfo{year}{2025}).
\newblock


\bibitem[Turki et~al\mbox{.}(2022)]%
        {mega-nerf}
\bibfield{author}{\bibinfo{person}{Haithem Turki}, \bibinfo{person}{Deva Ramanan}, {and} \bibinfo{person}{Mahadev Satyanarayanan}.} \bibinfo{year}{2022}\natexlab{}.
\newblock \showarticletitle{Mega-nerf: Scalable construction of large-scale nerfs for virtual fly-throughs}. In \bibinfo{booktitle}{\emph{IEEE/CVF Conference on Computer Vision and Pattern Recognition (CVPR)}}.
\newblock


\bibitem[Wei et~al\mbox{.}(2025)]%
        {lsgaussian}
\bibfield{author}{\bibinfo{person}{Linye Wei}, \bibinfo{person}{Jiajun Tang}, \bibinfo{person}{Fan Fei}, \bibinfo{person}{Boxin Shi}, \bibinfo{person}{Runsheng Wang}, {and} \bibinfo{person}{Meng Li}.} \bibinfo{year}{2025}\natexlab{}.
\newblock \showarticletitle{No Redundancy, No Stall: Lightweight Streaming 3D Gaussian Splatting for Real-time Rendering}. In \bibinfo{booktitle}{\emph{International Conference on Computer-Aided Design (ICCAD)}}.
\newblock


\bibitem[Xiong et~al\mbox{.}(2024)]%
        {gauuscene}
\bibfield{author}{\bibinfo{person}{Butian Xiong}, \bibinfo{person}{Zhuo Li}, {and} \bibinfo{person}{Zhen Li}.} \bibinfo{year}{2024}\natexlab{}.
\newblock \showarticletitle{GauU-Scene: A Scene Reconstruction Benchmark on Large Scale 3D Reconstruction Dataset Using Gaussian Splatting}.
\newblock \bibinfo{journal}{\emph{arXiv preprint arXiv:2401.14032}} (\bibinfo{year}{2024}).
\newblock


\bibitem[Ye et~al\mbox{.}(2024)]%
        {gsplat}
\bibfield{author}{\bibinfo{person}{Vickie Ye}, \bibinfo{person}{Ruilong Li}, \bibinfo{person}{Justin Kerr}, \bibinfo{person}{Matias Turkulainen}, \bibinfo{person}{Brent Yi}, \bibinfo{person}{Zhuoyang Pan}, \bibinfo{person}{Otto Seiskari}, \bibinfo{person}{Jianbo Ye}, \bibinfo{person}{Jeffrey Hu}, \bibinfo{person}{Matthew Tancik}, {and} \bibinfo{person}{Angjoo Kanazawa}.} \bibinfo{year}{2024}\natexlab{}.
\newblock \showarticletitle{gsplat: An Open-Source Library for Gaussian Splatting}.
\newblock \bibinfo{journal}{\emph{arXiv preprint arXiv:2409.06765}} (\bibinfo{year}{2024}).
\newblock


\bibitem[Ye et~al\mbox{.}(2025)]%
        {gbu}
\bibfield{author}{\bibinfo{person}{Zhifan Ye}, \bibinfo{person}{Yonggan Fu}, \bibinfo{person}{Jingqun Zhang}, \bibinfo{person}{Leshu Li}, \bibinfo{person}{Yongan Zhang}, \bibinfo{person}{Sixu Li}, \bibinfo{person}{Cheng Wan}, \bibinfo{person}{Chenxi Wan}, \bibinfo{person}{Chaojian Li}, \bibinfo{person}{Sreemanth Prathipati}, {and} \bibinfo{person}{Yingyan~Celine Lin}.} \bibinfo{year}{2025}\natexlab{}.
\newblock \showarticletitle{Gaussian Blending Unit: An Edge GPU Plug-in for Real-Time Gaussian-Based Rendering in AR/VR}. In \bibinfo{booktitle}{\emph{IEEE International Symposium on High Performance Computer Architecture (HPCA)}}.
\newblock


\bibitem[Yu et~al\mbox{.}(2022)]%
        {plenoxels}
\bibfield{author}{\bibinfo{person}{Alex Yu}, \bibinfo{person}{Sara Fridovich-Keil}, \bibinfo{person}{Matthew Tancik}, \bibinfo{person}{Qinhong Chen}, \bibinfo{person}{Benjamin Recht}, {and} \bibinfo{person}{Angjoo Kanazawa}.} \bibinfo{year}{2022}\natexlab{}.
\newblock \showarticletitle{Plenoxels: Radiance Fields without Neural Networks}. In \bibinfo{booktitle}{\emph{IEEE/CVF Conference on Computer Vision and Pattern Recognition (CVPR)}}.
\newblock


\bibitem[Zhai et~al\mbox{.}(2025)]%
        {splatloc}
\bibfield{author}{\bibinfo{person}{Hongjia Zhai}, \bibinfo{person}{Xiyu Zhang}, \bibinfo{person}{Boming Zhao}, \bibinfo{person}{Hai Li}, \bibinfo{person}{Yijia He}, \bibinfo{person}{Zhaopeng Cui}, \bibinfo{person}{Hujun Bao}, {and} \bibinfo{person}{Guofeng Zhang}.} \bibinfo{year}{2025}\natexlab{}.
\newblock \showarticletitle{SplatLoc: 3D Gaussian Splatting-based Visual Localization for Augmented Reality}.
\newblock \bibinfo{journal}{\emph{IEEE Transactions on Visualization and Computer Graphics (TVCG)}} (\bibinfo{year}{2025}).
\newblock


\bibitem[Zhao et~al\mbox{.}(2025)]%
        {grendel}
\bibfield{author}{\bibinfo{person}{Hexu Zhao}, \bibinfo{person}{Haoyang Weng}, \bibinfo{person}{Daohan Lu}, \bibinfo{person}{Ang Li}, \bibinfo{person}{Jinyang Li}, \bibinfo{person}{Aurojit Panda}, {and} \bibinfo{person}{Saining Xie}.} \bibinfo{year}{2025}\natexlab{}.
\newblock \showarticletitle{On Scaling Up 3D Gaussian Splatting Training}. In \bibinfo{booktitle}{\emph{International Conference on Learning Representations (ICLR)}}.
\newblock


\bibitem[Zoomers et~al\mbox{.}(2025)]%
        {progs}
\bibfield{author}{\bibinfo{person}{Brent Zoomers}, \bibinfo{person}{Maarten Wijnants}, \bibinfo{person}{Ivan Molenaers}, \bibinfo{person}{Joni Vanherck}, \bibinfo{person}{Jeroen Put}, \bibinfo{person}{Lode Jorissen}, {and} \bibinfo{person}{Nick Michiels}.} \bibinfo{year}{2025}\natexlab{}.
\newblock \showarticletitle{PRoGS: Progressive Rendering of Gaussian Splats}. In \bibinfo{booktitle}{\emph{IEEE/CVF Winter Conference on Applications of Computer Vision (WACV)}}.
\newblock


\end{thebibliography}

\end{document}